# A Comprehensive Review of Computer Vision in Sports: Open Issues, Future Trends and Research Directions

[1]Banoth Thulasya Naik[a], [2]Mohammad Farukh Hashmi[a], [3]Neeraj Dhanraj Bokde[b,*]

[a] Department of Electronics and Communication Engineering, National Institute of Technology, Warangal, India

[b] Department of Engineering - Renewable Energy and Thermodynamics, Aarhus University, 8000, Denmark

[1]thulasyramsingh@student.nitw.ac.in, [2]mdfarukh@nitw.ac.in, [3,]neerajdhanraj@mpe.au.dk

## Abstract:

Recent developments in video analysis of sports and computer vision techniques have achieved significant improvements to enable a variety of critical operations. To provide enhanced information, such as detailed complex analysis in sports like soccer, basketball, cricket, badminton, etc., studies have focused mainly on computer vision techniques employed to carry out different tasks. This paper presents a comprehensive review of sports video analysis for various applications high-level analysis such as detection and classification of players, tracking player or ball in sports and predicting the trajectories of player or ball, recognizing the teams strategies, classifying various events in sports. The paper further discusses published works in a variety of application-specific tasks related to sports and the present researchers views regarding them. Since there is a wide research scope in sports for deploying computer vision techniques in various sports, some of the publicly available datasets related to a particular sport have been provided. This work reviews a detailed discussion on some of the artificial intelligence (AI) applications in sports vision, GPU-based work stations, and embedded platforms. Finally, this review identifies the research directions, probable challenges, and future trends in the area of visual recognition in sports.

# 1 Introduction:

Automatic analysis of video in sports is a possible solution to the demands of fans and professionals for various kinds of information. Analyzing videos in sports has provided a wide range of applications, which include player positions, extraction of ball's trajectory, content extraction and indexing, summarization, detection of highlights, on demand 3D reconstruction, animations, generation of virtual view, editorial content creation, virtual content insertion, visualization and enhancement of content, game play analysis and evaluations, identifying player's actions, referee decisions and other fundamental elements required for the analysis of a game.

The task of player detection (identification) and tracking is very difficult because of many challenges, which include similar appearance of subjects, complex occlusions, unconstrained field environment, background, unpredictable movements, unstable camera motion, issues with calibration of low textured fields and the editing done to broadcast video, lower pixel resolution of players who are distant and smaller in the frame, motion blur etc. The simultaneous detection of players the ball and tracking them at once is quite challenging, because of the zigzag movements of the ball and player, change of ball from player to player, severe occlusion between players and ball, etc., and hence this study presents a survey of detection, classification, tracking, trajectory prediction and recognizing the team's strategies etc. in various sports. Detection and tracking of player is the only major requirement in some sports like cycling, swimming etc. As a result, as illustrated in Figure 1, this research classifies all sports into two categories: player-centered sports and ball-centered sports, with extensive analysis in Section 4.

Recent developments in video analysis of sports have a focus on the features of computer vision techniques which are used to perform certain operations for which these are assigned, such as detailed complex analysis like detection and classification of each player based on their team in every frame or by recognizing the jersey number to classify players based on their team will helps to classify various events where the player is involved. In higher level analysis, such as tracking the player or ball, many more such evaluations are to be considered for the evaluation of a player's skills, detecting the team's strategies, events and the formation of tactical positions such as mid field analysis in various sports like soccer, basketball etc., and also various sports vision applications such as smart assistants, virtual umpires, assistance coaches etc., have been discussed in Section 7. A higher level semantic interpretation is an effective substitute, especially in situations requiring real time analysis and minimal human intervention for exploitation of the delivered system outputs.

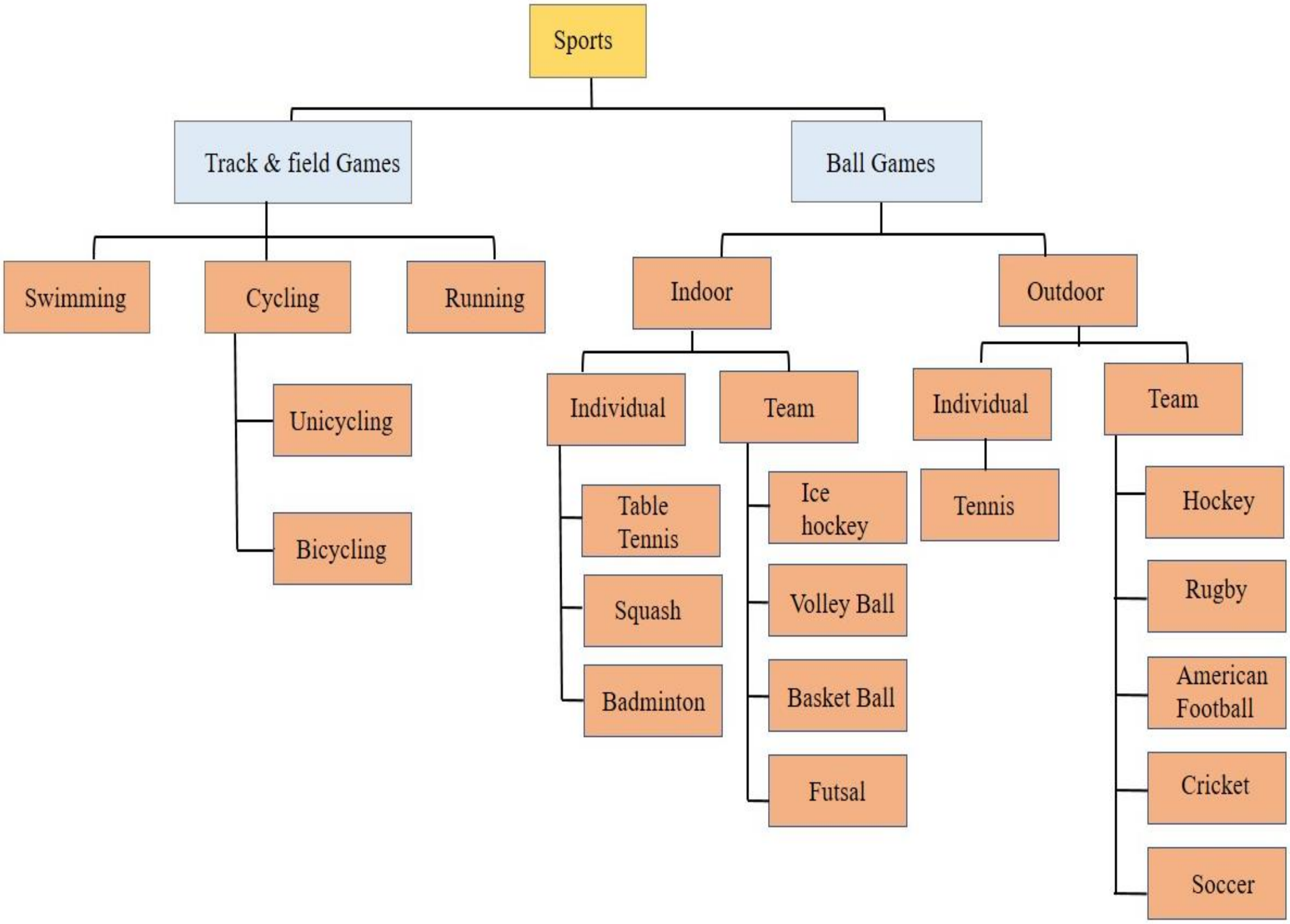

*Figure 1 Classification of different types of sports*

The main task of video summarization or highlight extraction is extracting key events of the game which provides users with an ability to view highlights as per their interests. For this purpose, it is required to detect, classify gestures, recognize the actions of referee/umpire, track players and the ball in key events like the time of goal scoring to analyze and classify different types of shots performed by players. The framework for processing and analyzing task-specific events in sports applications, such as playfield extraction, detection, and tracking of player/ball, etc. has been shown in figure 2, and detailed analyses of playfield extraction are discussed in Section 3.

A detailed review of research in the above-mentioned domains has been presented in this article and the data has compiled from papers which focus on computer vision based approaches that are used for each application, followed by inspecting key points and weaknesses, thereby investigating whether these methodologies in their current state of implementation can be utilized in real time sports video analysis systems.

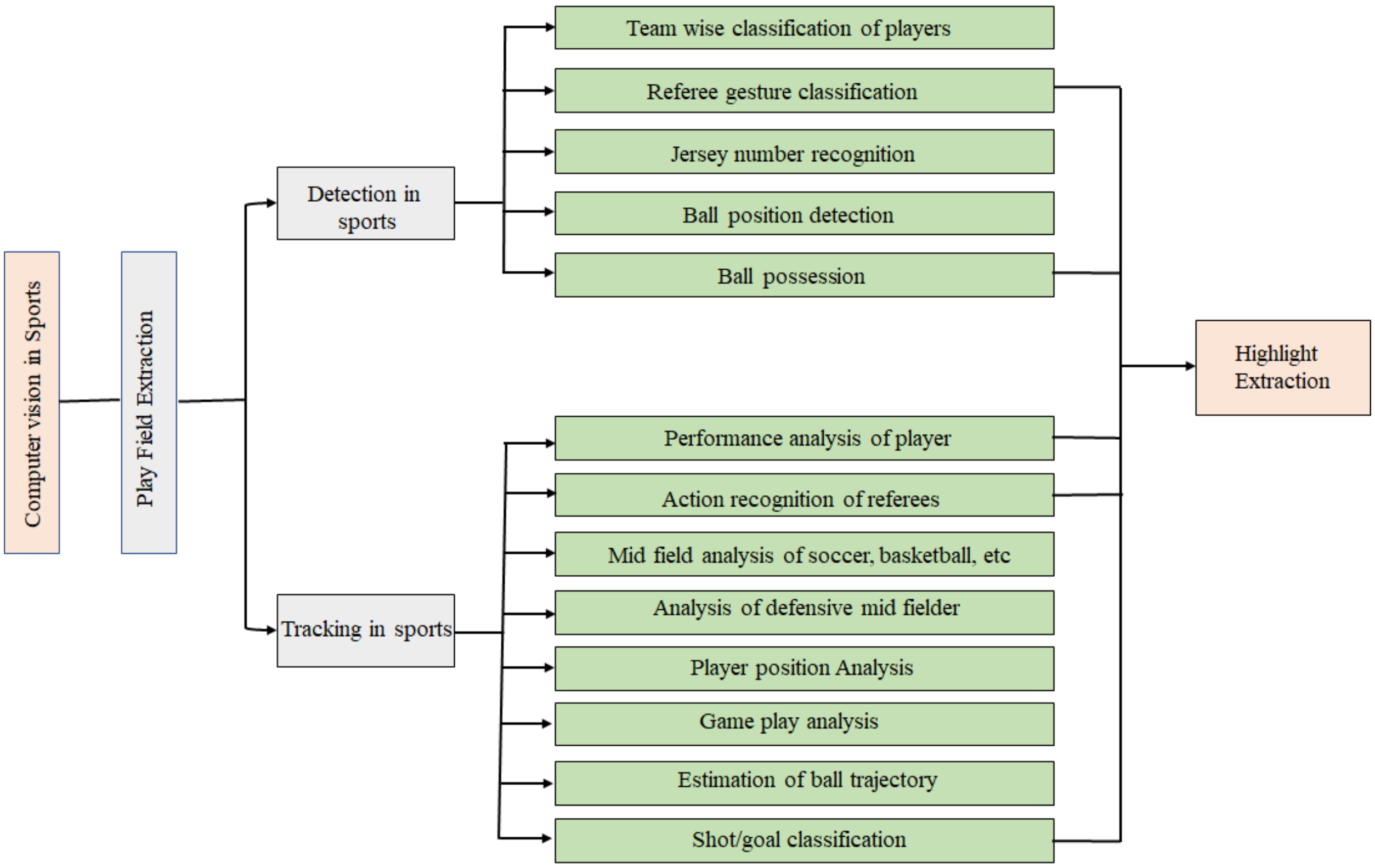

*Figure 2 Framework of processing and analysis of different applications in sports video*

## 1.1 Features of the Proposed Review

Some of the surveys and reviews published in different sports video processing and their main contributions are discussed and summarized in table 1 and listed below.

- D. Y. W. Tan et al. [1] researched on badminton movement analysis such as Badminton smashing, badminton service recognition, badminton swing and shuttle trajectory analysis.
- Robson P B et al. [2] presented a systematic review of sports data mining, which discusses the current panorama, themes, the dataset used, algorithms and research opportunities.
- N A Rahmad et al. [3] presented a survey on video based sports intelligence systems to recognize sports actions. They provided video based action recognition frameworks used in sports field and also discussed deep learning implementation in video based sports action recognition. They proposed a flexible method which classifies actions in different sports with different context and features as part of future research.

- Eline K et al. [4] presented an overview of 17 human motion capture systems which report the specs given by the manufacturer as well as calibration specs. This review helps researchers in the selection of a suitable motion capture system for experimental setups in various sports.
- M Manaffifard et al. [5] presented a survey on state-of-the-art (SOTA) algorithms for player tracking in soccer videos. They analyzed strengths and weaknesses of different approaches and presented the evaluation criteria for future research.
- Graham Thomas et al. [6] presented an analysis of computer vision based applications and research topics in sports field. They summarized some of the commercially available systems such as camera tracking and player tracking systems. They also incorporated some of the available datasets of different sports.
- Emily E C et al. [7] presented systematic review of literature on machine learning and deep learning for sports-specific movement recognition using inertial measurement unit and or computer vision data.
- PR Kamble et al. [8] presented an exhaustive survey of all the published research work on ball tracking in a categorical manner and also reviews the used techniques, their performance, advantages, and limitations with their suitability for a particular sport.
- HC Shih et al. [9] introduced the fundamentals of content analysis such as sports genre classification, and the overall status of sports video analytics. Also reviewed SOTA studies with prominent challenges identified in literature.
- Ryan Beal et al. [10] explored AI techniques that have been applied to challenges within a team sports such as match outcome prediction, tactical decision making, player investments, and injury prediction.
- Apostolidis E et al. [11] suggested a taxonomy of the existing algorithms and presented a systematic review of the relevant literature that shows the evolution of the deep learning based video summarization technologies.
- Yewande Adesida [289] explored a review to better understand the usage of wearable technology in sports to improve performance and avoid injury.
- Manju Rana [290] offered a thorough overview of the literature on the use of wearable inertial sensors for performance measurement in various sports.

**Table 1 Summary of Previous Survey and Reviews in different Sports**

| Ref | Hand crafted Algorithms | Machine Learning Algorithms | Sport and Application | | | | Discussed about Dataset | Aim of Review |
|---|---|---|---|---|---|---|---|---|
| | | | Sport | Detection | Tracking | Classification and | | |

| | | | | | | **Movement Recognition** | | |
|---|---|---|---|---|---|---|---|---|
| D. Y. W. Tan et al. [1] | ✓ | ✓ | Badminton | × | × | ✓ | × | Motion analysis |
| Robson P B et al. [2] | ✓ | ✓ | - | - | - | - | ✓ | Sports data mining |
| N A Rahmad et al. [3] | × | ✓ | - | ✓ | × | ✓ | ✓ | - |
| Eline K et al. [4] | ✓ | × | - | × | × | ✓ | × | Motion Capture |
| M Manaffifard et al. [5] | × | ✓ | Soccer | ✓ | ✓ | - | × | Player detection/tracking |
| Graham Thomas et al. [6] | × | ✓ | - | ✓ | ✓ | × | ✓ | Availability of datasets for sports |
| Emily E C et al. [7] | × | ✓ | - | × | × | ✓ | × | - |
| PR Kamble et al. [8] | ✓ | ✓ | Soccer | ✓ | ✓ | × | × | Ball Tracking |
| HC Shih et al. [9] | ✓ | ✓ | - | | × | | × | Content-Aware Analysis |
| Ryan Beal et al. [10] | × | ✓ | - | × | × | ✓ | × | - |
| Apostolidis E et al. [11] | × | ✓ | - | × | × | × | ✓ | Video Summarization |
| Yewande Adesida [289] | ✓ | x | - | - | - | - | × | Wearable technology in sports |
| Manju Rana [290] | ✓ | x | - | - | ✓ | ✓ | × | Wearable technology in sports |

The proposed survey mainly focuses on providing a proper and comprehensive survey of research carried out in computer vision based sports video analysis for various applications such as detection and classification of players, tracking player or ball and predicting the trajectories of player or ball, recognizing the team's strategies, classifying various events in sports field etc. and in particular, establishing a pathway for next-generation research in the sports domain. The features of this review are:

- In contrast to recently published review papers in the sports field, this article comprehensively reviews statistics of studies in various sports and various AI algorithms that have been used to cover various aspects viewed and verified in sports.
- It provides road map of various AI algorithms selection and evaluation criteria and also provided some of the publicly available datasets of different sports.
- It discusses various GPU-based embedded platforms for real time object detection and tracking framework to improve the performance and accuracy of edge devices.
- Besides, it demonstrates various applications in sports vision and possible research directions.

The rest of this paper is organized as follows. Section 2 provides statistical details of research in sports. Section 3 presents extraction data vis-a-vis various sports play fields, followed by a broader dimension that covers a wide range of sports and is reviewed in section 4. Some of the available datasets for various sports along and embedded platforms have been reviewed in section 5 and 6. Section 7 provides various application-specific tasks in the field of sports vision. Section 8 covers potential research directions, as well as different challenges to be overcome in sports studies. Last, but not the least, Section 9 concludes by describing the final considerations.

# 2 Statistics of Studies in Sports

Detection of the positions of the players at any given point of time is the basic step for tracking a player, which is also needed for graphics systems in sports for analysis and getting pictures of key moments in a game. Equipment and methods used in commercial broadcast analysis systems range from those depending on a manual operator clicking on the feet of the players with a calibrated camera image to an automated technique that involves segmentation to identify areas which likely correspond to players. For improvement of the performance of teams in sports like soccer, volleyball, hockey, badminton etc., analyzing the movements of players individually, and the real time formation of the teams, can provide a valuable real time insight for the coach of that team.

On obtaining a number of articles from various resources, the research articles are selected based on peer review, ranging from high impact factor online sources in the domain of player/ball/referee detection and tracking in sports, classification of objects in sports, behaviour and performance analysis of players, gesture recognition of referees/umpires, automatic highlight detection and score updating etc. Figure 3 provides overall information of sports research publications in past five years considered in this comprehensive survey article.

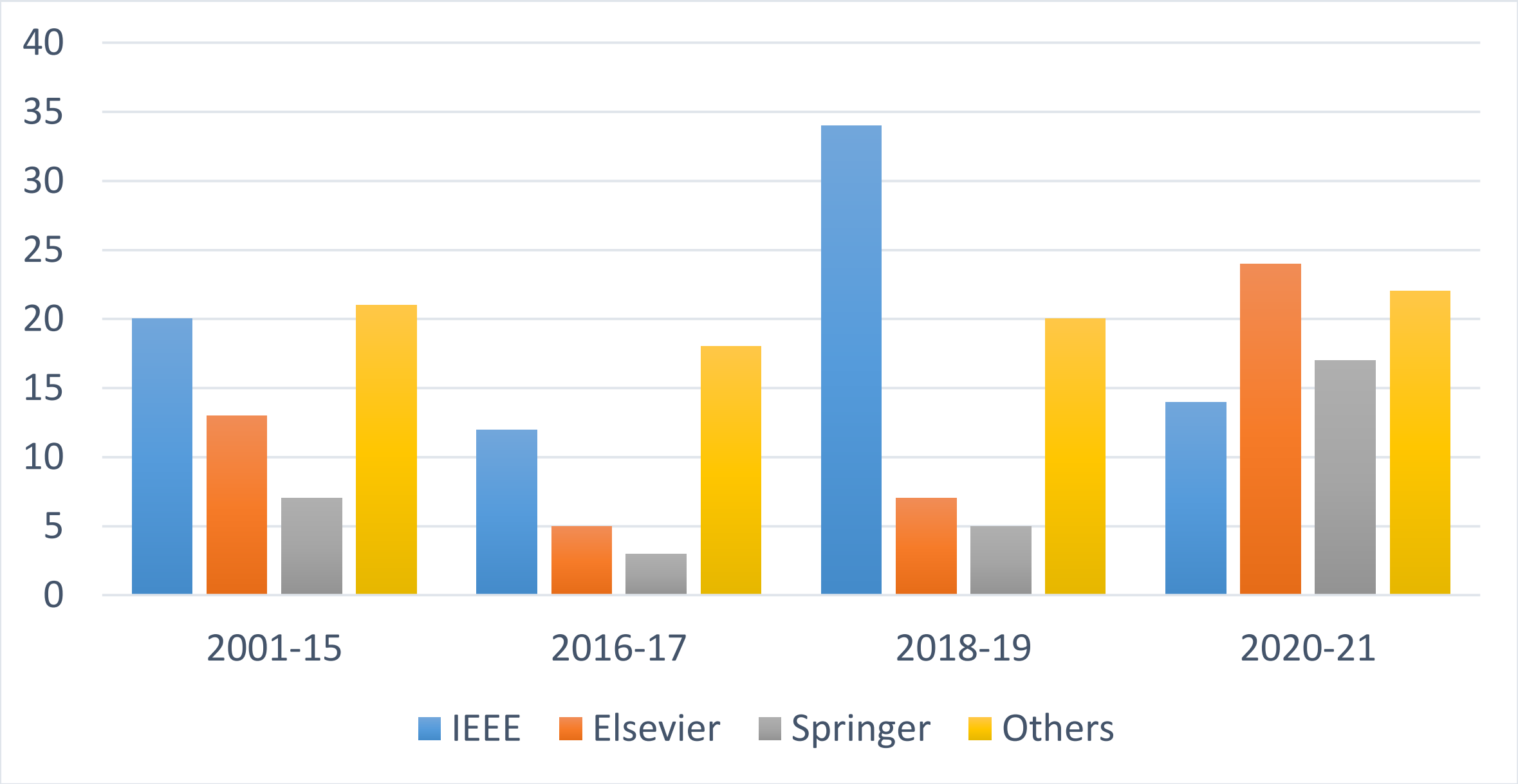

*Figure 3 Sports research progress in past five years*

Figure 4 provides the statistics of studies of various sports in various applications such as detecting/tracking the player and ball, trajectory prediction, classification, video summarization etc. which are published in various standard journals as presented in figure 3.

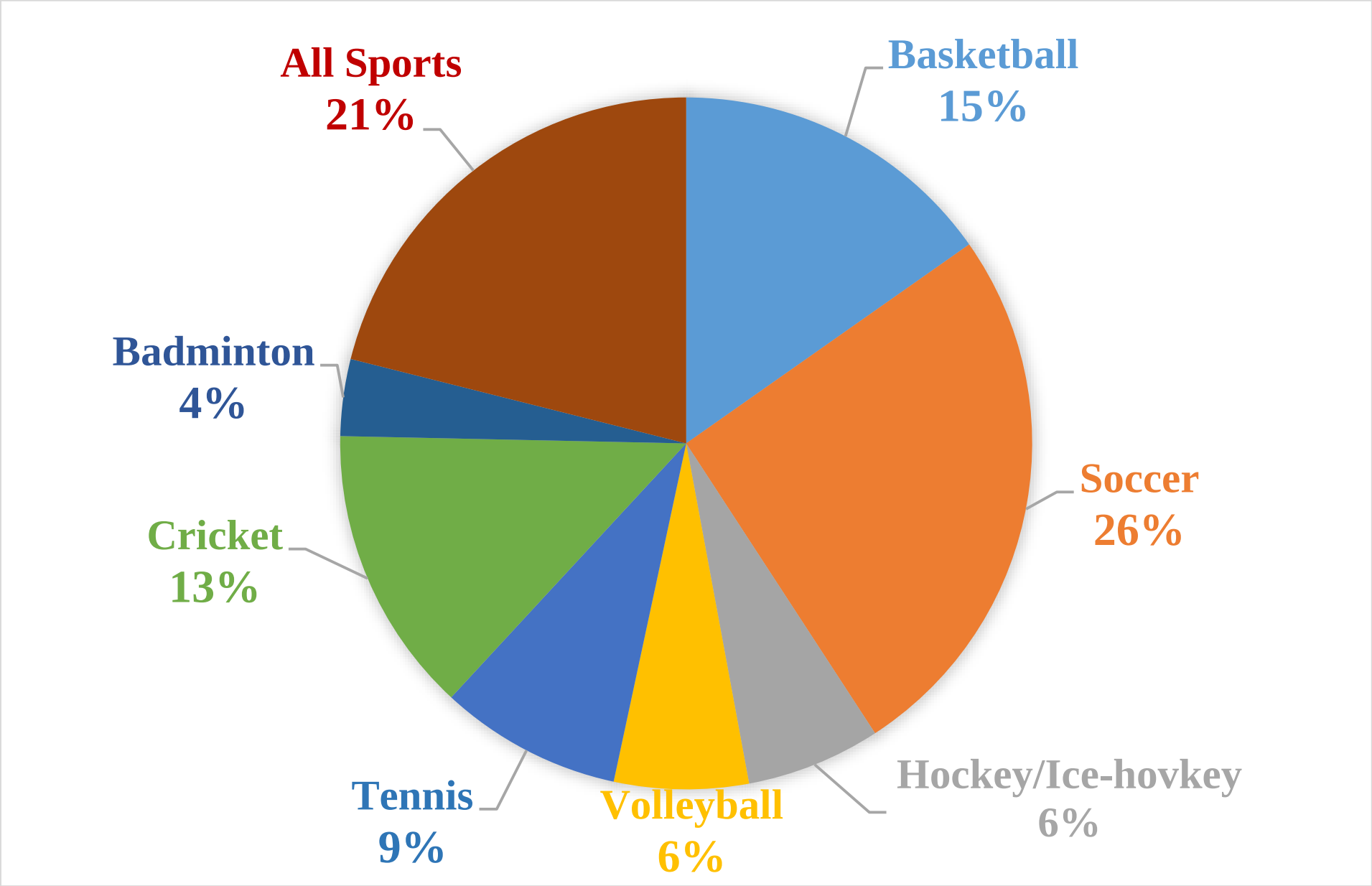

*Figure 4 Sports wise research progress*

# 3 Play Field Extraction in Various Sports

Detection of the sports field plays an important role in sports video analysis. Detection of playfield region has two objectives. One is to detect the playfield region from non-playfield areas as presented in [7], while the other is to identify primary objects from the background by filtering out redundant pixels such as grass, court lines. This provides a reduced pixel which requires processing and reduction of errors for simplifying player or ball detection and tracking phases, event extraction, pose detection etc. The challenges here include distinguishing the color of the playfield from that of the stadium, lighting conditions and sometimes weather, viewing angles and the shadows. Therefore, an accurate segmentation of the playfield cannot be achieved just by processing the color of the playfield under certain situations and making it constant without updating the statistics throughout the game. There is also an added noise when the player's dress matches that of the ground, and there appear shadows at the base of a player from different sources of light. Gaussian-based background subtraction technique [12] which is implemented using computer vision methods, generates the foreground mask as shown in figure 5.

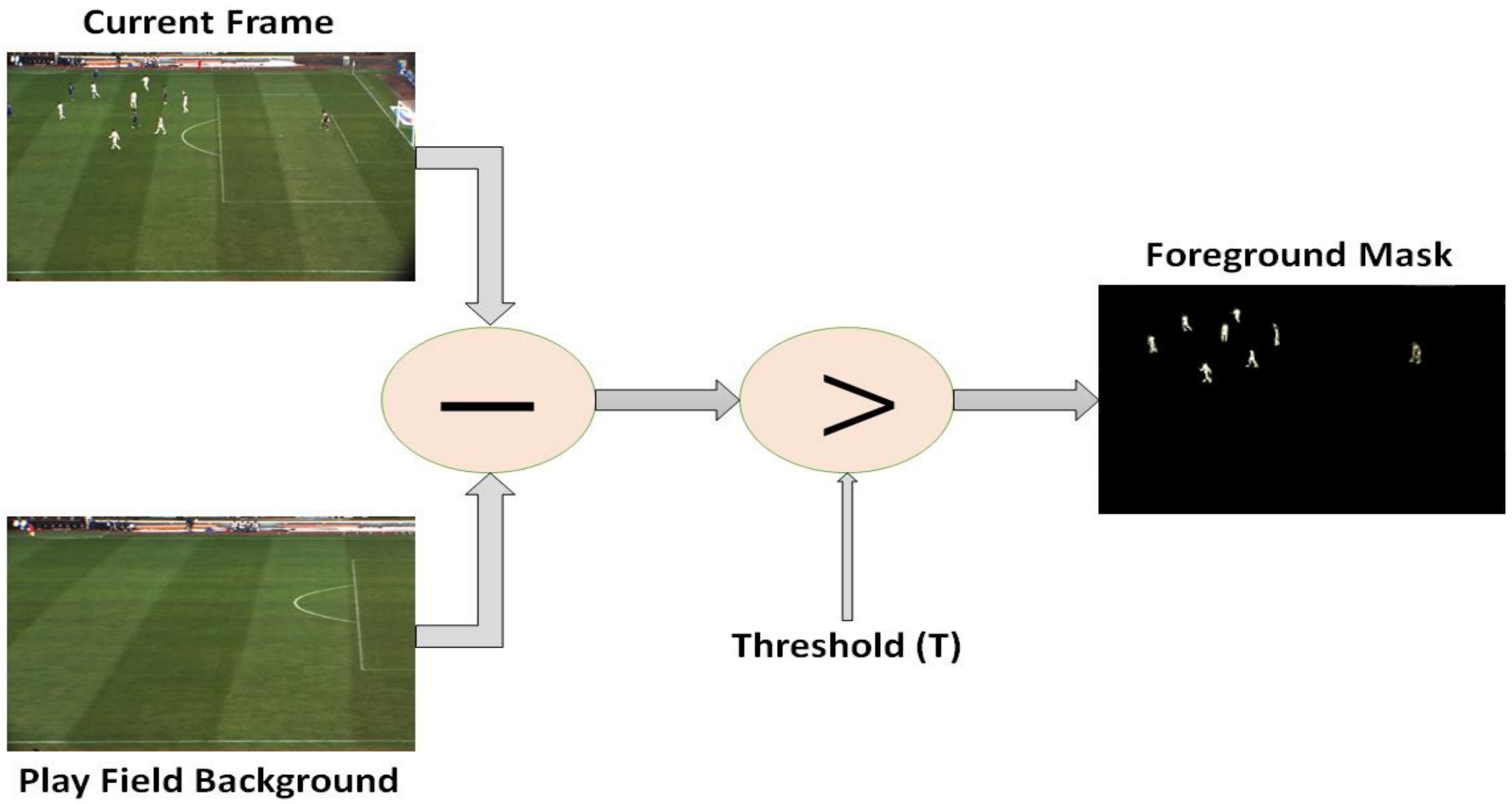

Figure 5 Background subtraction model

Researchers have used a single dominant color for detecting the playfield. Accordingly, some researchers have utilized the features of images in which illumination is not affected by transforming the images from RGB space to HIS [13, 14, and 15], YCbCr [16], normalized RGB [17, 18, 19].

For a precise capture of the movements of the players, tracking the ball and actions of referees, etc. on the field or court, it is necessary to calibrate the camera [5, 8] and also to use an appropriate number of cameras to cover the field. Though, some algorithms are capable of tracking the players, some other objects are also needed to be tracked in the dynamically complex situations of interest for detailed analysis of the events and extraction of the data of the subject of interest. Y Ohno et al. [20] presented an approach to extract the play field and track the players and ball using multiple cameras in soccer video. In [21, 22] presented an architecture which uses single (figure 6 (a)) and multi camera (figure 6 (c)) to capture a clear view of players and ball in various challenging and tricky situations such as severe occlusions, missing of the ball from the frames, etc. To estimate the players trajectory and team classification in [23, 24] presented a bird's eye view of the field to capture players precisely as shown in figure 6 (b). Various positions of the camera for capturing the entire field is presented in [25, 26] to detect and track the players/ball and estimating the position of the players etc.

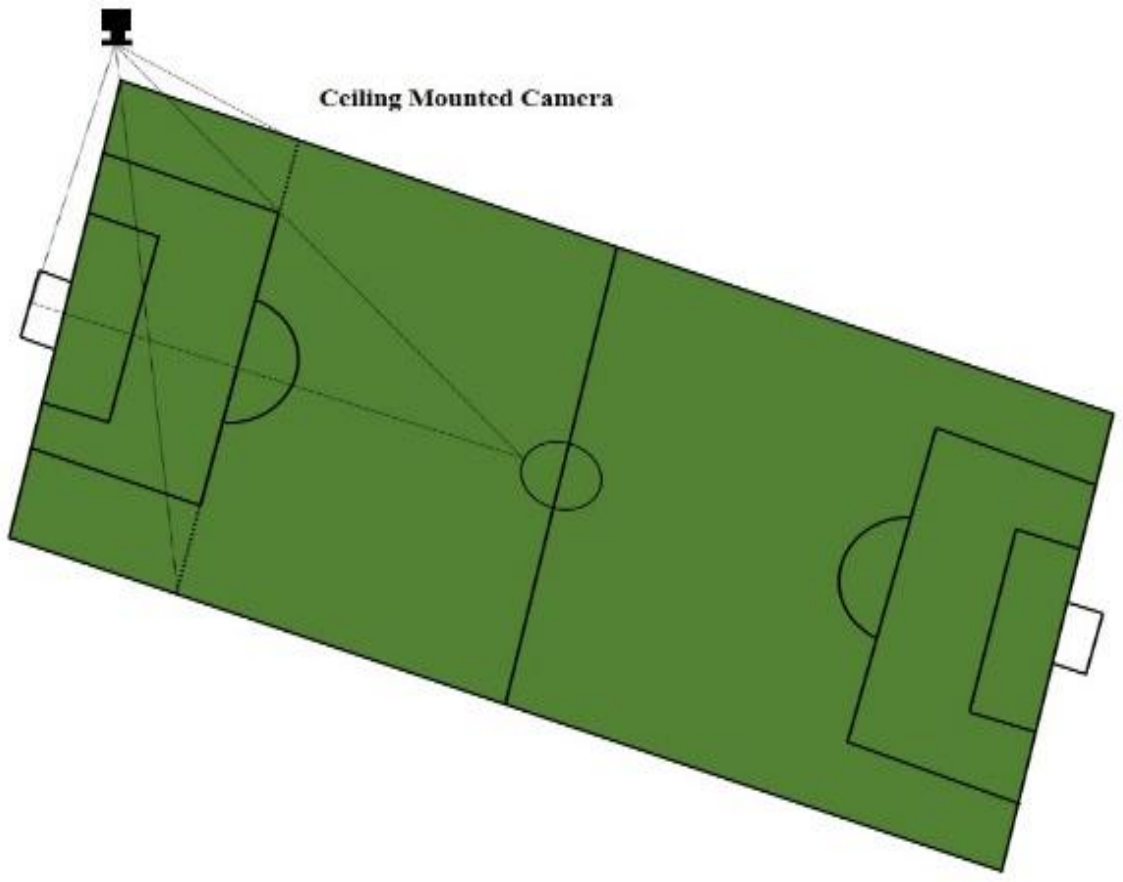

*Figure 6 (a) Ceiling Mounted Camera*

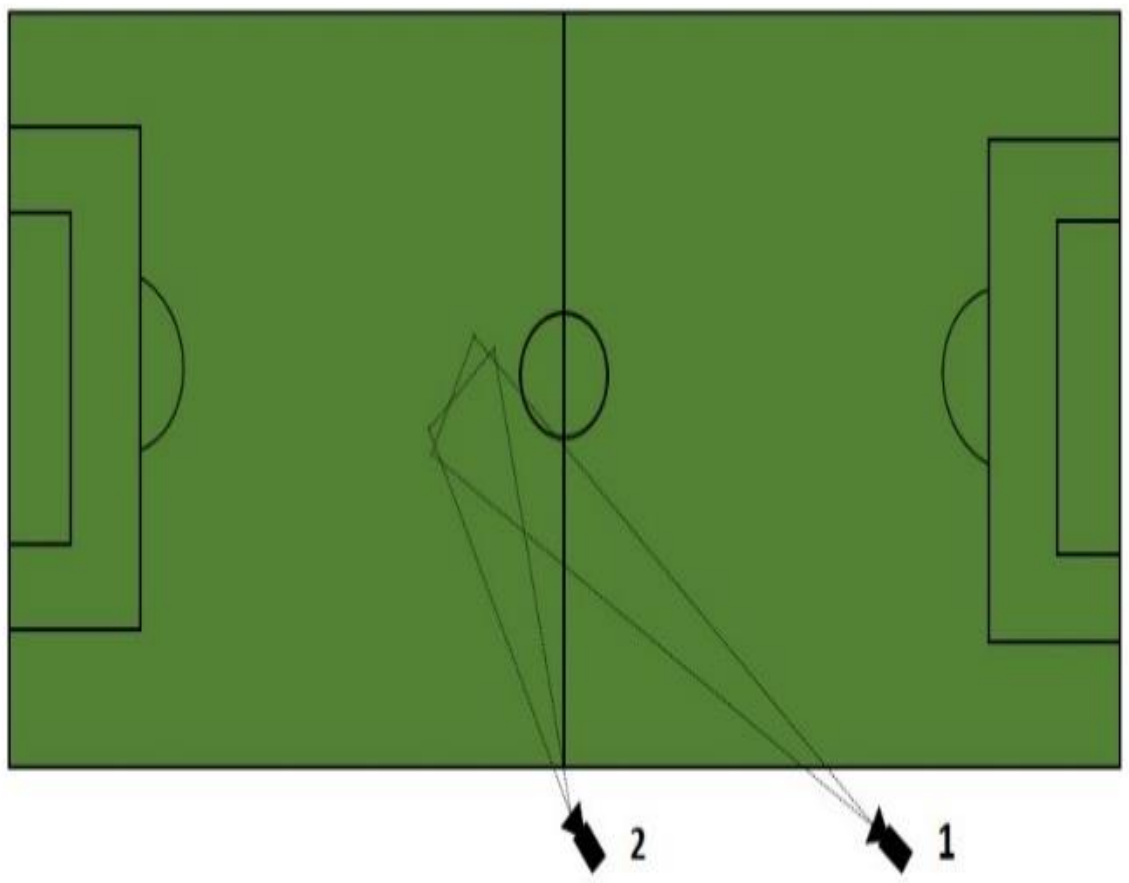

*Figure 6 (b) Birds eye view of Field*

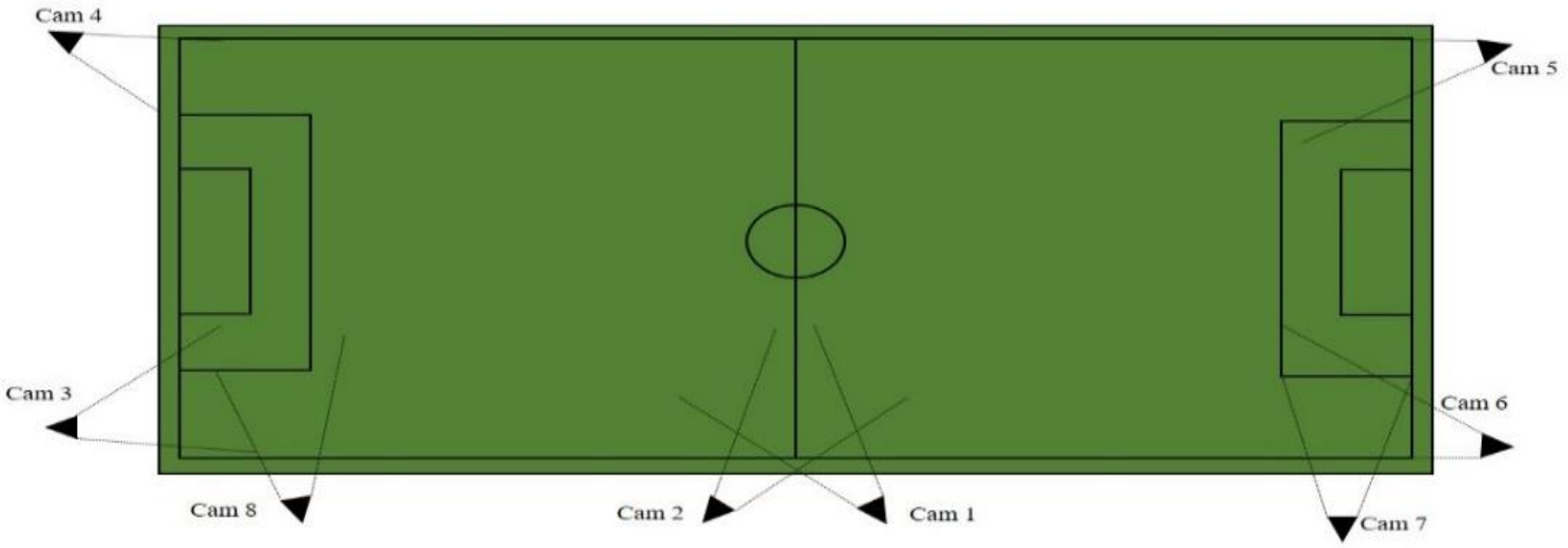

*Figure 6 (c) Multiple Cameras placed to cover complete playfield*

Morphological operations-based techniques can separate the playfield and non-playfield regions, but they cannot detect the lines in the playfield. The background subtraction based techniques generate foreground regions by subtracting the background frame from the current frame (i.e. by detecting moving objects in the frame), however they fails to detect the playfield lines as shown in figure 5. So, the best way to detect the playfield lines is by labeling the data as playfield lines (as shown in figure 7(a)), advertisements (as shown in figure 7(b)), and the non-playfield regions as shown in figure 7(c). Training the model using a dataset that is labeled as playfield lines, advertisements, and the non-playfield region as shown in figure 8 can detect and classify the playfield lines, advertisements, and the non-playfield region, which reduces the detection of false positives and false negatives.

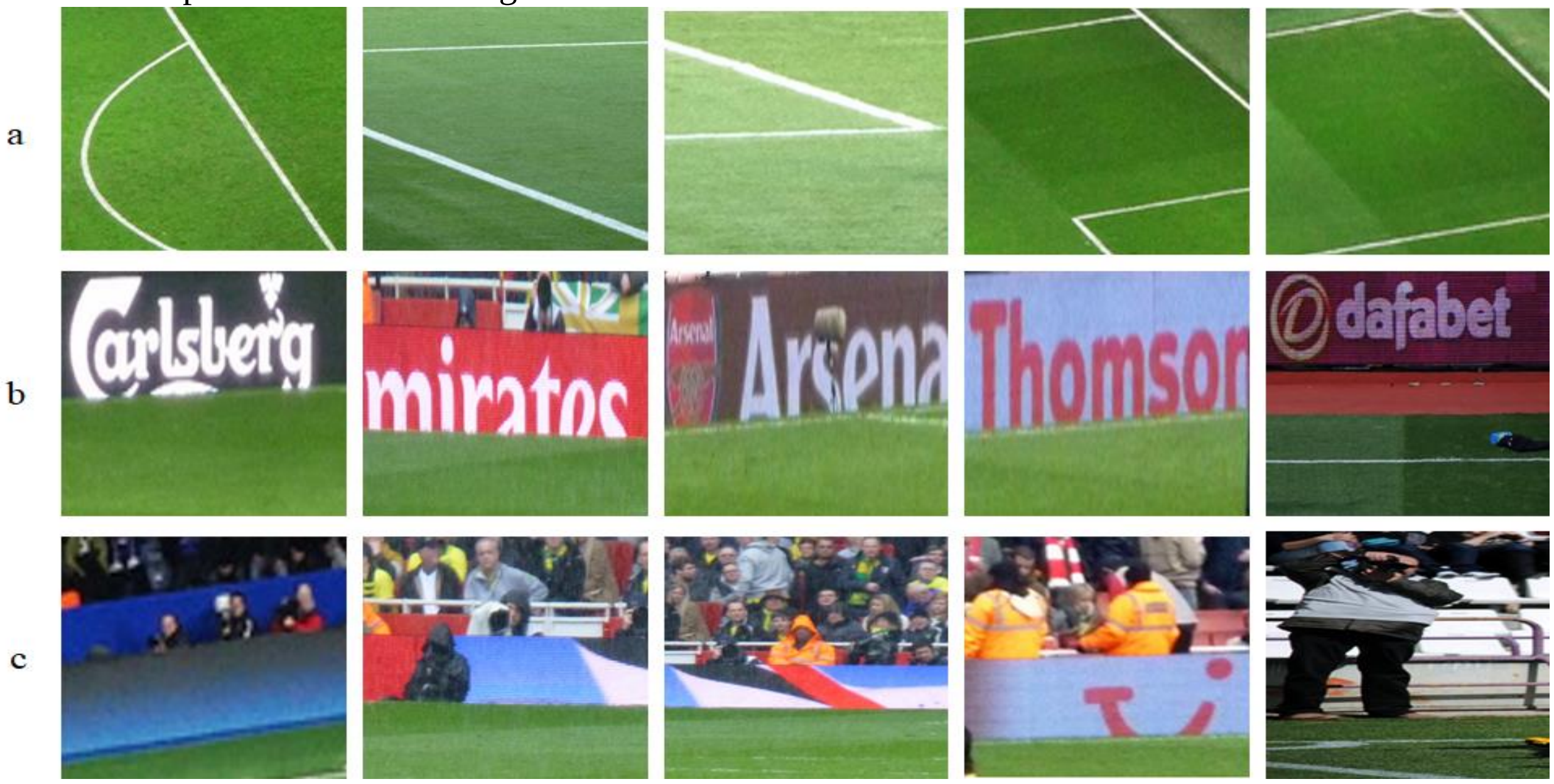

*Figure 7 Background labeled samples from the dataset a) Playfield lines b) Advertisements c) Non-playfield region.*

# 4 Literature Review

In this section, the overview of traditional computer-vision methods implemented for major application specifics in sports (such as detection, events classification/recognizing, tracking and trajectory prediction etc.) investigated by the researchers and their significant limitations has been discussed.

## 4.1 Basketball

Basketball is played by two teams of five players each. The object of the game is to score more points than opponent. The ball is passed, thrown, bounced, batted, or rolled from one player to another. Physical contact with an opponent can result in a foul if the contact impedes the desired movement of the player. The advent of computer vision techniques have effectively replaced manual analysis of tennis sports with fully automated systems. Recognizing the player action and classifying the events [27, 28, 29] in basketball videos helps to analyze the player performance. Player/ball detection and tracking in basketball videos is carried out in [30, 31, 32, 33, 34, 35] but fails in assigning specific identification to avoid the identity switching among the players when they crossed. By estimating the pose of the player, the trajectory of the ball [36, 37] is estimated from various distances to the basket. By recognizing and classifying the referee signals [38], player behaviour can be assessed and highlights of the game can be extracted [39]. The behaviour of a basketball team [40] can be characterized by the dynamics of space creation presented in [41, 42, 43, 44, 45, 46] that works to counteract space creation dynamics with defensive play presented in [47]. By detecting a specific location of the player and ball in the basketball court, the player movement can be predicted [48] and the ball trajectory [49, 50, 51] can be generated in three-dimensional which is a complicated task. It is also necessary to study the extraction of basketball players' shooting motion trajectory, combined with the image feature analysis method of basketball shooting, to reconstruct and quantitatively track the basketball players' shooting motion trajectory [52, 53, 54, 55]. However, it is difficult to analyze the game data for each play such as the ball tracking or motion of the players in the game, because the situation of the game changes rapidly, and the structure of the data should be complicated. Therefore, it is necessary to analyze the real time game play [56]. Table 2 summarizes various proposed methodologies to resolve various challenging tasks in basketball sport with their limitations.

**Table 2. Studies in Basketball**

| Ref. | Problem Statement | Proposed Methodology | Precision and Performance characteristics | Limitations and Remarks |
|---|---|---|---|---|
| L Long 2021 [29] | Recognizing action of basketball player by using image | Bi-LSTM Sequence2Sequence | The metrics used to evaluate the method are Spearman rank-order correlation coefficient, Kendall rank-order correlation coefficient, Pearson linear correlation | ➢ The methodology failed to recognize difficult actions due to which accuracy is reduced. |
| | recognition technique. | | coefficient, and Root Mean Squared Error and achieved 0.921, 0.803, 0.932, and 1.03 respectively. | ➢ The accuracy of action recognition can be improved with deep convolutional neural network. |
| Bertugli A et al. 2021 [51] | Multi-future trajectory prediction in basketball. | Conditional Variational Recurrent Neural Networks (RNN) - TrajNet++ | The proposed methodology was tested on Average Displacement Error and Final Displacement Error metrics. The methodology is robust if smaller the number and it has achieved 7.01 and 10.61. | ➢ Proposed methodology fail to predict the trajectories in case of uncertain and complex scenarios.<br>➢ As the behaviour of the basketball/payers are dynamic in nature, belief maps cannot steer future positions.<br>➢ Training the model with a dataset of different events can rectify the failures of predictions. |
| Y Wang et al. 2021 [56] | Predicting line-up performance of basketball player by analyzing the situation of the field. | RNN + NN | At the point guard (pg) position 4 candidates were taken and at the center (c) position 3 candidates were taken. The total score of pg candidates is 13.67, 12.96, 13.42, 10.39, and where the total score of c candidates is 10.21, 14.08, and 13.48 respectively. | - |
| Xubo Fu et al. 2020 [30] | Multiplayers tracking in basketball videos | YOLOv3 + Deep-SORT<br>Faster-RCNN + Deep-SORT<br>YOLOv3 + DeepMOT<br>Faster-RCNN + DeepMOT<br>JDE | Faster-RCNN provides better accuracy than YOLOv3 among baseline detectors.<br>Joint Detection and Embedding method performs better in accuracy of tracking and computing speed among multi object tracking methods. | ➢ Tracking in specific areas like severe occlusions and improving detection precision improves the accuracy and computation speed.<br>➢ By adopting frame extracting methods, in terms of speed and accuracy, it can achieve comprehensive performance, which may be an alternative solution. |
| Julius Žemgulys et al. 2020 [38] | Recognizing the referee signals from real time videos in basketball game. | HOG + SVM<br>LBP + SVM | Achieved an accuracy of 95.6% for referee signal recognition using local binary pattern features and SVM classification. | ➢ In case of noisy environment, significant chance of occlusion, unusual viewing angle and/or variability of gestures, the performance of the proposed method is not consistent.<br>➢ Detecting with jersey colour and eliminating all other detected elements in the frame can be the other solution to improve the accuracy of referee signal recognition. |
| L Wu et al. 2020 [28] | Event recognition in basketball videos | CNN | mAP for group activity recognition is 72.1% | ➢ Proposed model can recognize the global movement in the video.<br>➢ By recognizing the local movements the accuracy can be improved. |
| L Chen et al. 2020 [41] | Analyzing behaviour of the player. | CNN + RNN | Achieved an accuracy of 76.5% for four type of actions in basketball video. | ➢ Proposed model gives less accuracy for the actions like pass and foul.<br>➢ Also gives less accuracy of recognition and prediction on test dataset compared to valid dataset. |
| Y Yoon et al. 2019 [31] | Tracking ball movements and classification of players in basketball game | YOLO + Joy2019 | Jersey number recognition in terms of Precision achieved is 74.3%.<br>Player recognition in terms of Recall achieved 89.8%. | ➢ YOLO confuses the overlapped image for a single player. In the subsequent frame, the tracking ID of the overlapped player is exchanged, which causes wrong player information to be associated with the identified box. |
| L Wu et al. 2019 [27] | Events classifications in basketball videos | CNN + LSTM | Average accuracy using two-stage event classification scheme achieved 60.96%. | ➢ Performance can be improved by introducing information like individual player pose detection., player location detection etc. |
| C Tian et al. 2019 [47] | Classification of different defensive strategies of basketball payers, at particularly when they deviate from their initial defensive action. | KNN, Decision Trees and SVM | Achieved 69% of classification accuracy for automatic defensive strategy identification. | ➢ Considered only two defensive strategies “switch” and “trap” involved in Basketball.<br>➢ In addition, the alternative method of labeling large spatio-temporal datasets will also lead to better results.<br>➢ Future research may also consider other defensive strategies such as pick-and-roll and pick-and-pop. |
| Yu Z et al. 2018 [36] | Basketball trajectory prediction based on real data and generating new trajectory samples. | BLSTM + MDN | The proposed method performed well in terms of convergence rate and final AUC (91%), and proved deep learning models perform better than conventional models (GLM and GBM). | ➢ To improve the accuracy time series the prediction has to consider.<br>➢ By considering factors like player cooperation and defense when predicting NBA player positions, the performance of the model can be improved. |
| Zheng et al. 2017 [49] | Generating basketball trajectories. | GRU-CNN | Validated on hierarchical policy network (HPN) with ground truth and 3 baselines. | ➢ Proposed model failed in trajectory of three-dimensional basketball match. |
| W Liu et al. [2017] [39] | Score detection, highlight video generation basketball video. | BEI+CNN | Automatically analyses the basketball match, detects scoring and generates highlights. Achieved an accuracy, precision, recall, and f1-score of 94.59%, 96.55%, 92.31%, and 94.38%. | ➢ Proposed method is lacks in computation speed which achieved 5 frames per second.<br>➢ Therefore it cannot be implemented in a real-time basketball match. |
| V.Ramanathan et al. 2016 [32] | Multi-person event recognition in basketball video. | BLSTM | Event classification and event detection achieved in terms of mean average precision i.e. 51.6% and 43.5%. | ➢ High resolution dataset can improve the performance of the model. |
| K Wang et al. 2016 [42] | Player behaviour analysis. | RNN | Achieved an accuracy of 80% over offensive strategies. | ➢ The methodology fails in many factors such as complexity of interaction, distinctiveness, and diversity of the target classes and other extrinsic factors like reactions to defense, unexpected events such as fouls, and consistency of executions. |
| Rajiv C.S et al. 2016 [37] | Prediction of 3-point shot in basketball game | RNN | Evaluated in terms of AUC and achieved 84.30%. | ➢ Proposed method fails in case of high ball velocity and noisy nature of motion data. |

## 4.2 Soccer

Soccer is played using football and two teams of eleven players each compete to get the ball into the other team's goal, thereby scoring a goal. The players always confuse each other by changing their speed or direction unexpectedly. Due to their same jersey color, players look almost identical and are frequently involved in possessing the ball which leads to severe occlusions and tracking ambiguities. In such case, jersey number must be detected to recognize player [57]. Accurate tracking [58, 59, 60, 61, 62, 63, 64, 65, 66, 67, 291, 292] by detection [68, 69, 70, 71] of multiple soccer players as well as ball in real-time, is a major challenge to evaluate the performance of the players, to find their relative positions at regular intervals, and to link spatiotemporal data to extract trajectories. The systems which evaluate the player [72] or team performance [73] have the potential to reveal aspects of the game that are not obvious to the human eye. Such systems can successfully evaluate the activities of players [74] such as distance covered by players, shot detection [75, 76], number of sprints, player's position and their movements [77, 78], player's relative position concerning other players, possession [79] of the soccer ball and motion/gesture recognition of referee [80], predicting player trajectories for shot situations [81]. The generated data can evaluate individual player performance, occlusion handling [19] by detecting position of player [82], action recognition [83], predicting and classifying the passes [84, 85, 86], key event extraction [87, 88, 89, 90, 91, 92, 93, 94, 95, 96], tactical performance of the team [97, 98, 99, 100, 101], and analyzing the team tactics based on the team formation [102, 103, 104], generating highlights [105, 106, 107, 108]. Table 3 summarizes various proposed methodologies to resolve various challenging tasks in soccer sport with their limitations.

**Table 3 Studies in Soccer**

| Ref. | Problem Statement | Proposed Methodology | Precision and Performance characteristics | Limitations and Remarks |
|---|---|---|---|---|
| B. Thulasya Naik et al. 2022 [291] | Player and ball detection and tracking in soccer. | YOLOv3 and SORT | Methodology achieved tracking accuracy of 93.7% on multiple object tracking accuracy metrics with a detection speed of 23.7 FPS and a tracking speed of 11.3 FPS. | ➢ This methodology effectively handle's challenging situations, such as partial occlusions, players, and ball, reappearing after a few frames but fails when the players were severely occluded. |
| B. Thulasya Naik et al. 2022 [292] | Player, referee and ball detection and tracking by jersey color recognition in soccer. | DeepPlayer-Track | Model achieved a tracking accuracy of 96% and 60% on MOTA and GMOTA metrics respectively with a detection speed of 23 FPS. | ➢ The limitation of this method is that when the player with the same jersey color is occluded, the ID of the player is switched. |
| G Paul et al. 2021 [72] | Tracking soccer players to evaluate the number of goals scored by a player. | Machine Learning and Deep Reinforcement Learning. | Performance of player tracking model measured in terms of mAP and achieved 74.6%. | ➢ The method failed to track the ball at critical moments such as pass at beginning and shot.<br>➢ It also failed to overcome the identity switching problem. |
| H Cho et al. 2021 [89] | Extracting ball events to classify the player passing style. | Convolutional Auto-Encoder | Methodology evaluated in terms of accuracy and achieved 76.5% for 20 players. | ➢ Concatenation of auto-encoder and extreme learning machine techniques will improve the events classification performance. |
| K Ali et al. 2021 [96] | Detecting events in soccer sport. | Variational Auto-encoder and EfficientNet | Achieved F1-score of 95.2% event images and Recall of 51.2% on images not related to soccer at threshold value of 0.50. | ➢ Deep extreme learning machine technique which employs auto-encoder technique may enhance the event detection accuracy. |
| Anthony et al. 2020 [77] | Action spotting soccer video. | YOLO-like encoder | Algorithm achieved mAP of 62.5%. | - |
| S Kusmakar et al. 2020 [73] | Team performance analysis in soccer | SVM | Prediction models achieved an overall accuracy of 75:2% in predicting the correct segmental outcome of the likelihood of team making a successful attempt to score a goal on used dataset. | ➢ The proposed model failed in identifying the players that are more frequently involved in match states that end with an attempt at scoring i.e., a ''SHOT '' at goal which may assist sports analysts and team staff to develop strategies suited to an opponent's playing style. |
| Y Kim et al. 2020 [80] | Motion Recognition of assistant referees in soccer | AlexNet, VGGNet-16, ResNet-18, and DenseNet-121 | The proposed algorithm achieved 97.56% accuracy with real-time operations. | ➢ Though the proposed algorithm is immune to variations of illuminance caused by weather conditions, it failed in the case of occlusions between referee and players. |
| A Hassan et al. 2020 [101] | Predicting the attributes (Loss or Win) in soccer sport. | ANN | The proposed model predicts 83.3% for the win case and 72.7% for loss. | - |
| S Genki et al. 2019 [97] | Team tactics estimation in soccer videos. | DELM | Performance of the model are measured on precision, recall, and f1-score and achieved 87.6%, 88%, and 87.8%. | ➢ Team tactics are estimated on the basis of the relationship between tactics of two teams and ball possession.<br>➢ The method fails to estimate the team formation at the beginning of the game. |
| Y Ganesh et al. 2019 [83] | Action recognition in soccer | CNN based Gaussian Weighted event based Action Classifier architecture | Accuracy in terms of F1 Score is achieved 52.8% for 6 classes. | ➢ By classifying the actions into subtypes, accuracy of action recognition can be enhanced. |
| P.R Kamble et al. 2019 [59] | Detection and tracking of ball in soccer videos. | VGG – M CNN | Achieved an accuracy of 87.45%. | ➢ It could not detect when the ball moved out of play in the field, in the stands region, or from partial occlusion by players, or with ball color matching the player jersey. |
| K Zhang et al. 2019 [90] | Automatic event extraction for soccer videos based on multi-cameras. | YOLO | U-encoder is designed for feature extraction and has better performance in terms of accuracy compared with fixed feature extractor. | ➢ To carry out tactical analysis of the team, player trajectory needs to be analyzed. |
| Jackman et al. 2019 [75] | Shot detection in football game | MobileNetV2 | MobileNetV2 method performed better than other feature extractor methods. | ➢ By extracting the features with the MobileNetV2 and then using 3D convolution on the extracted features for each frame can improve the detection performance. |
| Lindstrom et al. 2019 [81] | Predicting player trajectories for shot situations | LSTM | Performance is measured in terms of f1-score and achieved 53%. | ➢ The model failed to predict the player trajectory in the case of players confused each other by changing their speed or direction unexpectedly. |
| Y Wu et al. 2018 [103] | Analyzing the team formation in soccer and formulating several design goals. | | The formation detection model achieved a max accuracy of 96.8%. | ➢ The model is limited to scalability as it cannot be operated on high resolution soccer videos.<br>➢ The results are bounded to a particular match, and it cannot evaluate the |
| | | OpenCV is used for back-end visualization. | | tactical schemes across different games.<br>➢ Visualization of real-time team formation is another drawback as it limits the visualization of non-trivial spatial information.<br>➢ By applying state-of-the-art tracking algorithms, one can predominantly improve the performance of tactics analysis. |
| Gerke et al. 2017 [57] | Player recognition with jersey number recognition. | Spatial Constellation + CNN | Achieved an accuracy of 82% by combining Spatial Constellation + CNN models. | ➢ The proposed model failed to handle the players that are not visible for certain periods.<br>➢ Predicting the position of invisible players could improve the quality of spatial constellation features. |
| S Chawla et al.2017 [84] | Evaluating and classifying the passes in football game. | SVM | The proposed model achieves an accuracy of 90.2% during a football match. | ➢ To determine the quality of each pass, some factors like pass execution of player at particular difficult situation, strategic value of pass, the riskiness of the pass etc. need to be included.<br>➢ To rate the passes in sequence, it is needed to consider the sequence of passes during which player is possessing the ball. |
| H Tepanyan 2017 [79] | Detecting the ball and predicting which team possess the ball. | Static RNN | Achieved test accuracy of 85.5%. | ➢ The proposed model is not robust and it does not detect the ball as it is small in size in far view frames. |
| S Genki et al. 2016 [98] | Team tactics estimation in soccer videos. | SVM | The performance of the methodology is measured in terms of precision, recall, and f1-score and achieved 98%, 97%, and 98%. | ➢ The model fails when audiovisual features could not recognize quick changes in the team tactics. |
| J brookes et al. 2016 [88] | Analyzing pass events in the case of non-obvious insights in soccer. | k-NN, SVM | To extract the features of pass location, they used heat map generation and achieved an accuracy of 87% in classification task. | ➢ By incorporating temporal information, the classification accuracy can be improved and it is also offers specific insights to situations. |
| Baysal et al. 2015 [58] | Tracking the players in soccer videos. | HOG + SVM | Player detection is evaluated in terms of accuracy and achieved 97.7%.<br>Classification accuracy using k-NN achieved 93% for 15 classes. | - |
| Moez B et al. 2010 [74] | Action classification in soccer videos | LSTM + RNN | The model achieves a classification rate of 92% on four types of activities. | ➢ By extracting the features of various activities, the accuracy of the classification rate can be improved. |

## 4.3 Cricket

In many aspects of cricket as well, computer vision techniques can effectively replace manual analysis. A cricket match has many observable elements including batting shots [109, 110, 111, 112, 113, 114, 115, 116], bowling performance [117, 118, 119, 120, 121, 122], number of runs or score depending on ball movement, detecting and estimating the trajectory of the ball [123], decision making on placement of players foot [128], outcome classification to generate commentary [124, 125], detecting umpire decision [126, 127]. Predicting an individual crickter's performance [129, 130] based upon his past record can be critical in the selection of team members in international competitions. This process is highly subjective and usually requires much expertise and negotiate decision making. By predicting the results of cricket matches [131, 132, 133, 134, 135] such as toss decision, home ground, player's fitness, player's performance criteria [136], and other dynamic strategies the winner can be estimated. The video summarization process provides a compact version of the original video to manage the interesting video contents easily. Also, the video summarization methods capture the viewer's interest by selecting exciting events from the original video [137, 138]. Table 4 summarizes various proposed methodologies with their limitations to resolve various application issues in cricket sport.

**Table 4 Studies in Cricket**

| **Ref.** | **Problem Statement** | **Proposed Methodology** | **Precision and Performance characteristics** | **Limitations and Remarks** |
|---|---|---|---|---|
| 2021 [112] | Shot classification in cricket. | CNN – Gated Recurrent Unit | It evaluated in terms of precision, recall, and f1-score and achieved 93.40%, 93.10%, and 93% for 10 types of shots. | ➢ By incorporating unorthodox shots which are played in t20 in dataset may improve the testing accuracy. |
| R Rahman et al. 2021 [120] | Detecting the action of the bowler in cricket. | VGG16 - CNN | It evaluated in terms of precision, recall, and f1-score and maximum average accuracy achieved is 98.6% for 13 classes. (13 types of bowling actions) | ➢ Training the model with the dataset of wrong actions can improve the detection accuracy. |
| 2021 [128] | Movement detection of the batsman in cricket. | Deep-LSTM | Model evaluated in terms of mean square error and achieved minimum error of 1.107. | - |
| 2021 [137, 138] | Cricket video summarization. | Gated Recurrent Neural Network + Hybrid Rotation Forest -Deep Belief Networks YOLO | The methodology evaluated in terms of precision, f1-score, accuracy and achieved 96.82%, 94.83% and 96.32% for four classes.<br>Yolo is evaluated on precision, recall, and f1-score and achieved 97.1%, 94.4%, and 95.7% for 8 classes. | ➢ Decision tree classifier performance is low due to the existence of a huge number of trees. Therefore, a small change in decision tree may improve the prediction accuracy.<br>➢ Extreme Learning Machines has faced the problem of over fitting, which can be overcome by removing duplicate data in dataset. |
| ChetanKapadiyaet al. 2020 [129] | Prediction of individual player performance in cricket | Efficient Machine Learning Techniques | Proposed algorithm achieves classification accuracy of 93.73% which is good compared with traditional classification algorithms. | ➢ Replacing machine learning techniques with deep learning techniques may improve the performance in prediction even in the case of different environmental conditions. |
| Md. Ferdouse Ahmed Foysal1 et al. 2019 [109] | Classification of different batting shots in cricket. | CNN | Average classification in terms of precession is 0.80, Recall is 0.79 and F1-score is 0.79. | ➢ To improve the accuracy of classification, deep learning algorithm has to be replaced with a better neural network. |
| RohitKumaret al. 2019 [124] | Outcome classification task to create automatic commentary generation. | CNN + LSTM | Maximum of 85% of training accuracy and 74% validation accuracy | ➢ Due to the unavailability of standard dataset for ball by ball outcome classification in cricket, the accuracy is not up to mark. Also better accuracy leads to automatic commentary generation in sports. |
| Md. Kowsheret al. 2019 [126] | Detecting third umpire decision and Automated scoring system in cricket game. | CNN + Inception V3 | It holds 94% accuracy in Deep Conventional Neural Network (DCNN) and 100% in Inception V3 for the classification of umpire signal in order to automate scoring system of cricket. | ➢ To build automated umpiring system based on computer vision application and artificial intelligence, the results obtained in this paper are more enough. |
| MdNafeeAl Islam et al. 2019 [117] | Classification of cricket bowlers based on their bowling actions. | CNN | The test set accuracy of the model is 93.3% which demonstrates its classification ability. | ➢ The model lacks data for detecting spin bowlers. As the dataset is confined to left arm bowlers, the model misclassifies the right arm bowlers. |
| Muhammad Zeeshan Khan et al. 2018 [110] | Recognition of various batting | Deep - CNN | The proposed models are able to recognize | ➢ As the model is dependent on frame per second of the video, it fails to recognize |
| | shots in cricket game | | a shot being played with 90% accuracy. | when the frame per second increases. |
| PushkarShukla et al. 2018 [125] | Automatic highlight generation in the game of cricket. | CNN + SVM | Mean Average Precision of 72.31% | ➢ The proposed method cannot clear metrics to evaluate the false positives in highlights. |
| Aravind Ravi et al. 2018 [127] | Umpire poses detection and classification in cricket. | SVM | VGG19-Fc2 Player testing accuracy = 78.21% | ➢ Classification and summarization technique can minimize false positives and false negatives. |
| AFTAB KHAN et al. 2017 [111] | Activity recognition for quality assessment of batting shots. | Decision Trees, k-Nearest Neighbours, and Support Vector Machines. | The proposed method identifies 20 classes of batting shots with an average F1-score of 88% based on recorded movement of data. | ➢ In order to assess the player's batting caliber, certain aspects of batting also need to be considered i.e. position of batsman before playing shot and way of batting shots for a particular bowling types can be modeled. |
| 2016 [131] [132] | Predicting the outcome of the cricket match. | k-NN, Naïve Bayesian, SVM, and Random Forest | Achieved an accuracy of 71% upon the statistics of 366 matches. | ➢ Imbalance in the dataset is one of the causes to get less accuracy. ➢ Deep learning methodologies may give promising results by training with a dataset that included added features. |
| D. Bhattacharjee et a. 2012 [119] | Performance analysis of the bowler. | Multiple regression mode | Variation in ball speed has a feeble significance in influencing the bowling performance (p-value being 0.069). The variance ratio of the regression equation to that of the residuals (F-value) is given by 3.394 with corresponding p-value 0.015. | - |
| R.I.Subramanian et al. 2009 [130] | Predicting performance of the player. | Multilayer perception Neural Network | The model achieves an accuracy of 77% on batting performance and 63% on bowling performance. | - |

## 4.4 Tennis

Tennis is one of the most popular sports across the globe. A meticulous analysis of the game is needed to reduce human errors and extract several statistics from the visual feed of the game. Automated ball and player tracking is one such class of systems which requires sophisticated algorithms for analysis.

The primary data for tennis is obtained from the ball and player tracking systems, such as HawkEye[139, 140] and TennisSense [26, 141]. The data from this systems can be used to detect and track the ball/player [142, 143, 144, 145], visualizing the overall tennis match [146, 147] and predicting trajectories of ball landing positions [148, 149, 150], player activity recognition [151, 152, 153], analyzing the movements of the player and ball [154], analyzing the player behaviour [155] and predicting the next shot movement [156], real time tennis swing classification [157]. Table 5 summarizes various proposed methodologies to resolve various challenging tasks in tennis sport with their limitations.

**Table 5 Studies in Tennis**

| Ref. | Problem Statement | Proposed Methodology | Precision and Performance characteristics | Limitations and Remarks |
|---|---|---|---|---|
| G Wu et al. 2021 [140] | Monitoring and Analyzing tactics of tennis player. | YOLOv3 | Model achieved mAP of 90% with 13 FPS on high resolution images. | ➢ Using lightweight backbone for detection module can improve the processing speed. |
| N Bai et al. 2021 [153] | Player action recognition in tennis sport. | Temporal Deep Belief Network (Unsupervised Learning Model) | Accuracy of recognition rate is 94.72% | ➢ If two different movements are similar then the model is failed to recognize the actual action. |
| M Kevin et al. 2021 [157] | Tennis swing classification. | SVM, Neural Network (NN), K-NN, Random Forest, Decision Tree | Maximum classification accuracy of 99.72% achieved using NN with a Recall of 1.<br>Second highest classification accuracy of 99.44% was achieved using K-NN with a Recall of 0.98. | ➢ If the play styles of the players are different but the patterns are the same, in that case models failed to classify the actual swing direction. |
| J Cai et al. 2020 [151] | Player activity recognition in tennis game. | LSTM | The average accuracy of player activity recognition based on the historical LSTM model was 0.95, and that of the typical LSTM model was 0.70. | ➢ The model lacks real-time learning ability and requires much computing time at the training stage.<br>➢ The model is also lacks in online learning ability. |
| B Giles et al. 2019 [142] | Automatic detection and classification of change of direction from player tracking data in tennis game. | Random Forest Algorithm | Among all the proposed methods, model 1 had the highest F1-score of 0.801, as well as the smallest rate of false negative classification (3.4%) and average accuracy of 80.2% | ➢ In case of non-linear regression analysis, classification performance of the proposed model is not up to mark. |
| T Fernando et al. 2019 [148] | Prediction of shot location and type of shot in tennis game. | Generative Adversarial Network (GAN) (Semi-Supervised Model) | Performance factor is measured based on minimum distance recorded between predicted and ground truth shot location. | ➢ Performance of the model deviates for the different player styles as it trained on limited player dataset. |
| Tom Polk et al. 2019 [154] | Analyzing individual tennis match by capturing spatio temporal data player and ball movements. | For data extraction, player and ball tracking system such as HawkEye is used. | Generation of 1-D space charts for patterns and point outcomes to analyze the player activity. | ➢ Performance of the model deviates for different matches as it was trained only on limited tennis matches. |
| S.V Mora et al. 2017 [152] | Action recognition in tennis | 3 Layered LSTM | The classification accuracies as follows:<br>I. Improves from 84.10% to 88.16% for players of mixed abilities.<br>II. Improves from 81.23% to 84.33% for amateurs and from 87.82% to 89.42% for professionals.<br>When trained using the entire dataset. | ➢ The detection accuracy can be increased by incorporating spatio-temporal data and combining the action recognition data with statistical data. |
| Xinyu Wei et al. 2016 [156] | Shot prediction and player behavior analysis in tennis | For data extraction, player and ball tracking system such as HawkEye is used and Dynamic Bayesian Network for shot prediction. | By combining factors (Outside, Left Top, Right Top, Right Bottom) together, speed, start location, player movement achieved better results of 74% AUC. | ➢ As the model is trained on limited data (only elite players), it cannot be performed on ordinary players across multiple tournaments. |
| Xiangzeng Zhou et al. 2015 [143] | Ball tracking in tennis | Two-Layered Data Association | Evaluation results in terms of precision, recall, F1-score are of 84.39%, 75.81%, 79.87% for Australian open tennis match and 82.34%, 67.01%, 73.89% for U.S open tennis match. | ➢ The proposed method cannot handle multi object tracking and furthermore it is possible to integrate audio information to facilitate making a high level analysis of game. |
| Guangyu Zhu et al. 2007 [155] | Highlight extraction from rocket sports videos based on human behavior analysis. | SVM | The proposed algorithm achieved an accuracy of 90.7% for tennis videos and 87.6% for badminton videos. | ➢ The proposed algorithm fails to recognize player, as the player is a deformable object of which the limbs make free movement during action recognition. |

## 4.5 Volleyball

Volleyball is a team sport in which two teams of six players are separated by a net. Each team tries to score points by grounding a ball on the other team's court under organized rules. So, detecting and analyzing the player activities [158, 159, 160], detecting play patterns and classifying tactical behavior's [161, 162, 163, 164], predicting league standings [165], detecting and classifying spiking skills [166, 167], estimating the pose of the player [168], tracking player [169], tracking the ball [170] etc., are the major aspects of volleyball sport. Predicting the ball trajectory [171] in volleyball game by observing the motion of setter player. Table 6 summarizes various proposed methodologies to resolve various challenging tasks in volleyball sport with their limitations.

**Table 6 Studies in Volleyball**

| Ref. | Problem Statement | Proposed Methodology | Precision and Performance characteristics | Limitations and Remarks |
|---|---|---|---|---|
| T Haritha et al. 2021 [168] | Group activity recognition by tracking players. | CNN + Bi-LSTM | Model achieved an accuracy of 93.9%. | ➢ Model fails to track the players if video is taken from dynamic camera.<br>➢ Temporal action localization can improve the accuracy of tracking the players in severe occlusion condition. |
| Z Kai et al. 2021 [169] | Recognizing and classifying player's behaviour. | SVM | Achieved recognition rate of 98% for correct samples 349. | - |
| Sebastian Wenninger,et al. 2020 [162] | Classification of tactical behaviors in beach volleyball. | RNN + GRU | The model achieves better classification results as prediction accuracies ranges from 37% for the forecasting the attack and direction to 60% of the prediction of success. | ➢ By employing state-of-the-art method and training on proper dataset which has continuous positional data, it is possible predict tactic behavior and set/match outcomes. |
| Y Tian 2020 [170] | Motion estimation the sport of volleyball | Machine Vision and Classical particle filter. | Tracking accuracy is 89% | ➢ Replacing with deep learning algorithms gives better results. |
| Fasih Haider1 et al. 2019 [163] | Assessing the use of Inertial Measurement Units in the recognition of different volleyball actions. | KNN, Naïve Bayes, SVM | Unweighted Average Recall of 86.87% | ➢ By incorporating different frequency domain features, the performance factor can be improved. |
| ShuyaSuda et al. 2019 [171] | .Predicting the ball trajectory in volleyball game by observing the | Neural Network | The proposed method predicts 0.3 s in advance of the trajectory of the volleyball based on the motion of setter player. | ➢ In case of predicting the 3D body position data, it records a large error. This can be overcome by training a |
| | motion of setter player. | | | properly annotated large data on state-of-art-methods. |
| Thomas Kautz et al. 2017 [159] | Activity recognition in beach-volleyball game | Deep Convolutional LSTM | The approach achieved classification accuracy of 83.2%, which is superior compared with other classification algorithms. | ➢ Instead of using wearable devices, the computer vision architectures can be used to classify the activities of the players in volleyball. |
| J Wei et al. 2021 [165] | Volleyball skills and tactics analysis | ANN | Evaluated in terms of Average Relative Error for 10 samples and achieved 0.69%. | - |
| Mostafa S. Ibrahim et al.2016 [160] | Group activity recognition in volleyball game | LSTM | Group activity recognition of accuracy of the proposed model in volleyball is 51.1%. | ➢ The performance of architecture is poor because of lack of hierarchical considerations of individual and group activity dataset. |

## 4.6 Hockey/Ice-hockey

Field hockey, also called hockey, is an outdoor game played by two opposing teams of 11 players each who use sticks curved at the striking end to hit a small, hard ball into their opponent's goal. It is called field hockey to distinguish it from the similar game played on ice. So, detecting [172] and tracking player/hockey ball, recognizing the actions of player [173, 174, 175], estimating the pose of the player [176], classifying and tracking the players of the same team or of different teams [177], referee gesture analysis [178, 179], hockey ball trajectory estimation etc. are the major aspects of hockey sport.

Whereas ice hockey is the game between two teams, each usually having six players, who wear skates and compete on an ice rink. The object is to propel a vulcanized rubber disk, the puck, past a goal line and into a net guarded by a goaltender. With its speed and its frequent physical contact, ice hockey has become one of the most popular of international sports. So, detecting/tracking the player [180, 181, 182], estimating the pose of the player [183], classifying and tracking with different identification the players of the same team or of different teams, tracking the ice hockey puck [184], classification of puck possession events [185] etc., are the major aspects of the ice-hockey sport. Table 7 summarizes various proposed methodologies to resolve various challenging tasks in hockey/ice-hockey sport with their limitations.

**Table 7 Studies in Hockey**

| **Ref.** | **Problem Statement** | **Proposed Methodology** | **Precision and Performance characteristics** | **Limitations and Remarks** |
|---|---|---|---|---|
| Ș Melike et al. 2021 [172] | Detecting the player in hockey. | SVM, Faster R-CNN SSD and YOLO | HD+SVM achieved best results in terms of accuracy, recall and f1-score of 77.24%, 69.23%, and 73.02%. | ➢ Model is failed to detect the players in occlusion conditions. |
| V Kanav et al. 2021 [184] | Localizing puck Position and Event recognition. | Faster RCNN | Evaluated in terms of AUC and achieved 73.1%. | ➢ By replacing the detection method with YOLO series can improve the performance. |
| Alvin Chan et al. 2020 [177] | Identification of player in hockey. | ResNet + LSTM | Achieves player identification accuracy of over 87% on split dataset. | ➢ Some of the jersey number classes such as 1 to 4 are incorrectly predicted. The diagonal numbers from 1 to 100 are falsely classified due to the small amount of training examples. |
| KeerthanaRangasamy et al. 2020 [173] | Activity recognition in hockey game. | LSTM | The proposed model recognizes the activities like free hit, goal, penalty corner, long corner with an accuracy of 98%.<br>. | ➢ As the proposed model is focusses on the spatial features, it does not recognize activities like free hit and long corner as they appear as similar patterns.<br>➢ By including temporal features and incorporating LSTM to the model, the model is robust to performance accuracy. |
| Zixi Cai et al. 2019 [176] | Pose estimation and temporal based action recognition in hockey sport. | VGG19 +LiteFlowNet + CNN | A novel approach is designed and achieved an accuracy of 85% for action recognition. | ➢ The architecture is not robust to abrupt changes in video, i.e. it fails to predict hockey stick.<br>➢ Activities like goal scored, or puck location are not recognized. |
| KanavVat set al. 2019 [183] | Action recognition in ice hockey using player pose sequence. | CNN +LSTM | Performance of the model is better in similar classes like passing and shooting. It achieved 90% of parameter reduction and<br>80% of floating point reduction on HARPET dataset. | ➢ As the number of hidden units to LSTM increases, the number of parameters also increases which leads to overfitting, and low test accuracy. |
| Konstantin S et al. 2018 [174] | Human activity recognition in hockey sport. | CNN +LSTM | F1-score of 67% is calculated for action recognition on multi-labeled imbalanced dataset. | ➢ Performance of the model is poor because of improper imbalanced dataset |
| MehrnazFani et al. 2017 [175] | Player action recognition in ice hockey game | CNN | Accuracy of the actions recognized in hockey game is 65% and when similar actions are merged accuracy rises to 78%. | ➢ Pose estimation problems due to severe occlusions while motions blur due to speed of the game and also due to lack of proper dataset to train models, causing low accuracy. |

## 4.7 Badminton

Badminton is one of the most popular racket sports which includes tactics, techniques, and precise execution movements. To improve the performance of the player, technologies play a key role in optimizing training of players which determines the movements of the player [186] during training and game situations such as action recognition [187, 188, 189], analyzing performance of player [190], detecting and tracking shuttlecock [191, 192, 193] etc. Table 8 summarizes various proposed methodologies to resolve various challenging tasks in badminton sport with their limitations.

**Table 8 Studies in in Badminton**

| **Ref.** | **Problem Statement** | **Proposed Methodology** | **Precision and Performance characteristics** | **Limitations and Remarks** |
|---|---|---|---|---|
| Z Cao et al. 2020 [191] | Shuttlecock detection problem of a badminton robot. | Tiny YOLOv2 and YOLOv3 | Results show that comparing with state-of-art methods, proposed networks achieved good accuracy with efficient computation. | ➢ Proposed method fails to detect in different environmental conditions.<br>➢ As it uses the binocular camera to detect 2D shuttlecock, it cannot detect the 3D shuttlecock trajectory. |
| N A Rahmad et al. 2020 [187] | Automated badminton player action recognition in badminton game. | AlexNet+CNN, GoogleNet+CNN and SVM | Recognition of badminton actions by the linear SVM classifier for both AlexNet and GoogleNet using local and global extractor method is 82% and 85.7%. | ➢ The architecture can be improved fine tuning end-to-end manner with a larger dataset on feature extracted at different fully connected layers. |
| Teem steels et al. 2020 [188] | Badminton activity recognition | CNN | Nine different activities have been distinguished: seven badminton strokes, displacement and moments of rest. With accelerometer data, accurate estimation has been made using CNN with 86% precision. Accuracy raised to 99% when gyroscope data is combined with accelerometer data. | ➢ Computer vision techniques can be employed instead of sensors. |
| N A Rahamad et al. 2019 [189] | Classification of badminton matches images to recognize the different actions done by the athletes. | AlexNet, GoogleNet, Vgg-19 + CNN | Significantly, GoogleNet model has the highest accuracy compared to other models in which only two hit actions were falsely classified as non-hit action. | ➢ The proposed method classifies the actions of hit and non-hit and it can be improved by classifying more actions in various sports. |
| W Chen et al. 2019 [192] | Tracking shuttlecock in badminton sport | AdaBoost Algorithm which can be trained using OpenCV Library. | The performance of the proposed algorithm was evaluated based on precision and it achieved an average precision accuracy of 94.52% with 10.65 fps. | ➢ The accuracy of tracking shuttlecock is enhanced by replacing state-of-the-art AI algorithms. |
| Kokum W et al. 2017 [186] | Tactical movement classification in badminton | KNN | The average accuracy of player position detection is 96.03% and 97.09% on two halves of a badminton court. | ➢ The unique properties of application such as the length of frequent trajectories or the dimensions of the vector space may improve classification performance. |

## 4.8 Miscellaneous

Player detection and tracking is the major requirement in athletic sports like running, swimming [194, 195], and cycling. In sports like table tennis [196], squash [197, 198], golf [199] etc. ball detection and tracking, player pose detection [200] are the challenging tasks. In ball-centric sports like rugby, American football, handball, baseball etc. ball/player detection [201–208] and tracking [209-219], analyzing the action of player [220-226], events detection and classification [227-231], performance analysis of player [232-234], referee identification and gesture recognition, etc. are the major challenging tasks. Video highlight generation is a subclass of video summarization [235-238] which may be viewed as a subclass of sports video analysis. Table 9 summarizes various proposed methodologies to resolve various challenging tasks in various sports with their limitations.

**Table 9 Studies in various sports**

| **Ref.** | **Problem Statement** | **Proposed Methodology** | **Precision and Performance characteristics** | **Limitations and Remarks** |
|---|---|---|---|---|
| Liu Wei 2021 [208] | Beach sports image recognition and classification. | CNN | The model achieved recognition accuracy of 91%. | ➢ Lightweight networks of deep learning algorithms can improve the recognition accuracy and can also be implemented in real-time scenario. |
| Y Cao 2020 [195] | Motion image segmentation in the sport of swimming | GDA + SVM | The performance of Symmetric Difference Algorithm is measured in terms of recall and it achieved 76.2%. | ➢ Using advanced optimization techniques such as Cosine Annealing Scheduler with deep learning algorithms may improve the performance. |
| Hegazy H et al. 2020 [196] | Identifying and recognizing wrong strokes in table tennis sport. | k-NN,<br>SVM,<br>Naïve Bayes | Performs various ML algorithms and achieves an accuracy of 69.93% using Naïve Bayes algorithm. | ➢ A standard dataset can improve the accuracy of recognizing the wrong strokes in table tennis. |
| Ruiheng Zhang et al. 2020 [209] | Multi-player tracking in sports | Cascade Mask RCNN | The proposed Deep Player Identification method studies the patterns of jersey number, team class, and pose-guided partial feature. In order to handle player identity switch, the method correlates the coefficients of player ID in the K-shortest path with ID. The proposed framework achieves the state-of-art performance. | ➢ When compared with existing methods, computation cost is higher and can be considered a major drawback of the proposed framework.<br>➢ To refine 2D detection, temporal information needs to be considered and can be transferred to tracking against real-time performance like soccer, basketball etc. |
| R L Castro et al. 2019 [212] | Individual player tracking in sports events. | Deep Neural Network | Achieved an Area Under Curve (AUC) of 66% | ➢ Tracking by jersey number recognition may increase the performance of the model. |
| Cao Zhi-chao et al. 2019 [200] | Skelton-based key pose recognition and classification in sports | Boltzmann machine + CNN<br>Deep Boltzmann machine + RNN | The proposed architecture successfully analyses feature extraction, motion attitude model, motion detection and behavior recognition of sports posture. | ➢ The architecture is bound to individual-oriented sports and can be further implemented on group based sports, in case of challenges like severe occlusion, misdetection due to failure in blob detection in object tracking. |
| Mohammad Ashraf Russo et al. 2019 [222] | Human action recognition and classification in sports | VGG 16 + RNN | The proposed method achieved an accuracy of 92% for ten types of sports classification. | ➢ The model fails in the case of scaling up the dataset for larger classification which shows ambiguity between players and similar environmental conditions.<br>➢ Football, Hockey; Tennis, Badminton; Skiing, Snowboarding; these pairs of |
| | | | | classes have similar environmental feature's, thus it's only possible to separate them based on relevant actions which can be achieved by state-of-the-art methods. |
| Ali Javed et al. 2019 [236] | Replay and key event detection for sports video summarization | Extreme Learning Machine (ELM) | The framework is evaluated on dataset which consists of 20 videos of four different sports categories. It achieves an average accuracy of 95.8% which illustrates the significance of the method in terms of key-event and replay detection for video summarization. | ➢ Performance of the proposed method drops in case of absence of gradual transition of replay segments.<br>➢ It can be extended by incorporating artificial intelligence techniques. |
| SubhajitChaudhury et al. 2019 [227] | Event detection in sports videos for unstructured environments with arbitrary camera angles. | Mask RCNN + LSTM | The proposed method is accurate in unsupervised player extraction, which is used for precise temporal segmentation of rally scenes. It is also robust to noise in the form of camera shaking and occlusions. | ➢ It can be extended to double games with fine-grained action recognition for detecting various kinds of shots in an unstructured videos and it can be extended to analyze videos of games like cricket, soccer etc. |
| Yedong Li et al. 2019 [228] | Human motion quality assessment in complex motion scenarios. | 3 Dimensional CNN | . Achieved an accuracy of 81% on MS-COCO dataset. | ➢ Instead of the Stochastic Gradient Descent technique for learning rate, using the Cosine annealing scheduler technique may improve the performance. |
| Stephen Karungaru et al. 2019 [210] | Court detection using markers, player detection and tracking using a drone. | Template Matching + Particle Filter | The proposed method achieves better accuracy (94%) in case of overlapping of two players | ➢ As the overlapping of players increases the accuracy of detection, tracking decreases due to similar features of same team players.<br>➢ The method uses template matching algorithm, which can be replaced with deep learning based state-of-art algorithm to acquire better results. |
| QiuliHui. 2019 [211] | Target tracking theory and analyses its advantages in video tracking. | Mean Shift + Particle Filter | Achieves better tracking accuracy compared to existing algorithms like TMS and CMS algorithms. | ➢ If the target scales changes then the tracking of player fail's due to the unchanged window of mean-shift algorithm. Also it cannot track objects which are |
| | | | | similar to the background color.<br>➢ The accuracy of tracking players can be improved by replacing them with artificial intelligence algorithms. |
| Antonio Tejero-de-Pablos et al. 2018 [235] | Automatically generating a summary of sports video. | 2D CNN + LSTM | Describes a novel method for automatic summarization of user-generated sports videos and demonstrated the results for Japanese fencing videos. | ➢ The architecture can be improved by fine tuning end-to-end manner with a larger dataset for illustrating potential performance and also to evaluate in the context of wider variety of sports. |
| Ke Xu et al. 2017 [226] | Action Recognition and classification | SVM | Achieved an accuracy of 59.47% on HMDB 51 dataset. | ➢ In case, the object takes up most part of the frame, the human detector cannot completely cover the body of the object. This leads to missing movements of body parts like hands and arms. And also recognition of similar movements are the challenges for this architecture. |

## 4.9 Overview of Machine Learning/Deep Learning Techniques

There are multiple ways to classify, detect, and track objects to analyze the semantic levels involved in various sports. It paves the way for player localization, jersey number recognition, event classification, trajectory forecasting of the ball, etc., in a sports video with a much better interpretation of an image as a whole.

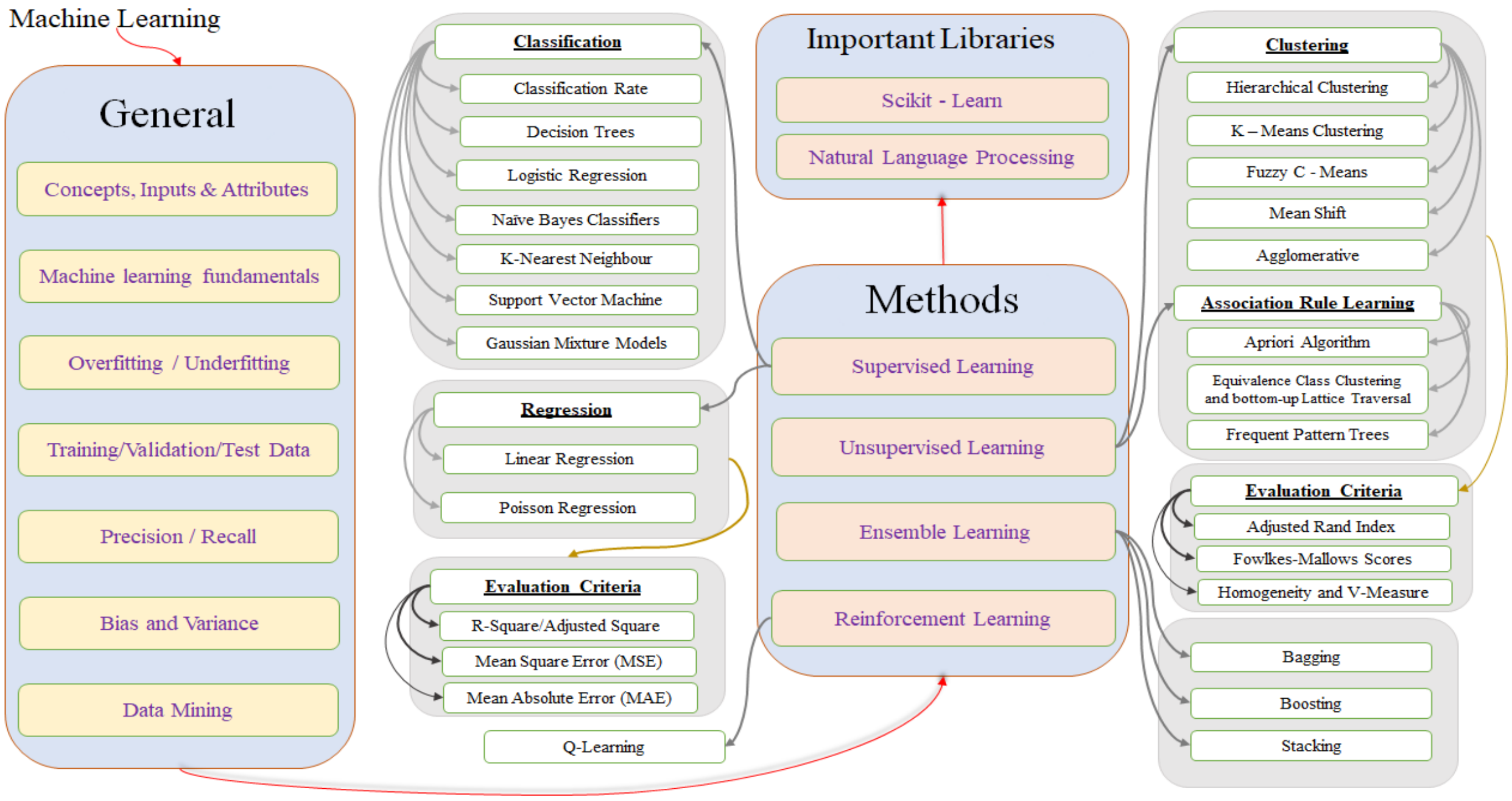

Figure 8: Block diagram of road map to machine learning architecture selection and training

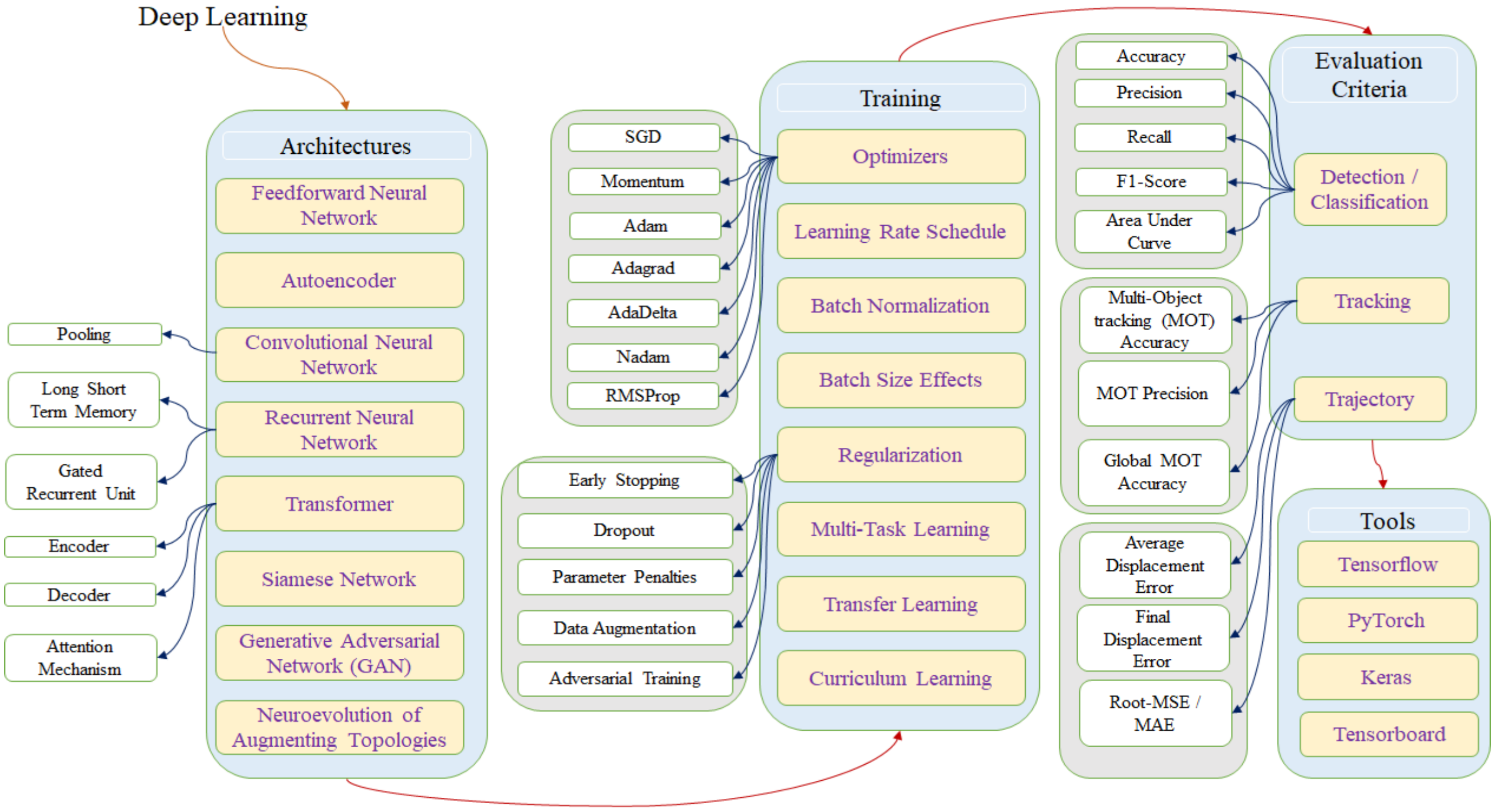

Figure 9: Block diagram of road map to deep learning architecture selection and training

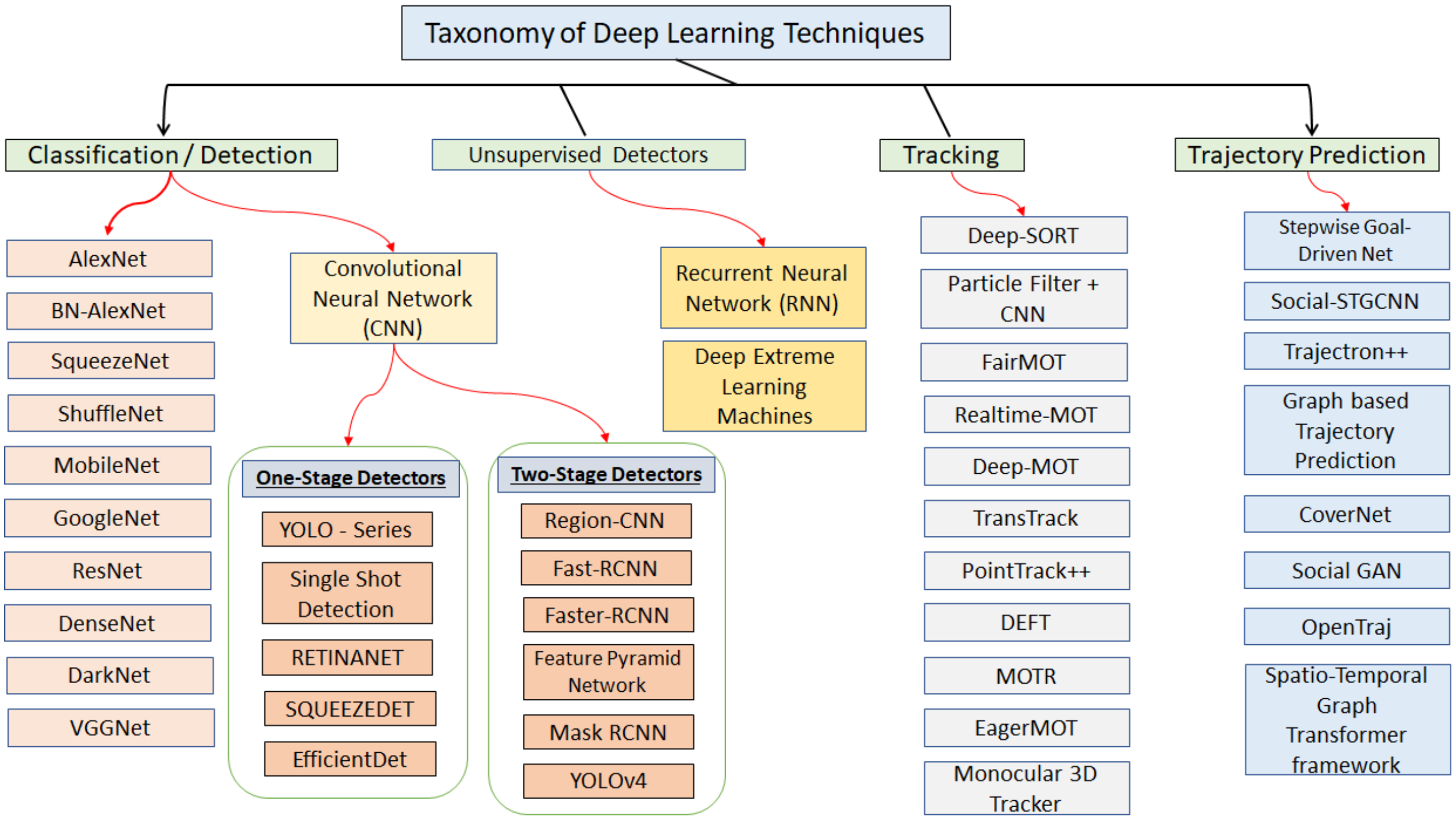

Figure 10: Overview of Deep Learning Algorithms of classification/detection, tracking and trajectory prediction

The selected AI algorithm is better if it is tested and benchmarked on the different data. For that to evaluate the robustness of AI algorithms some metrics are required, which measures the performance of particular AI algorithm to enable better selection. Figure 8 depicts the road map of the machine learning algorithms' general information, methods, and evaluation criteria for a particular task and required libraries/tools for training the model. Whereas Figure 9 depicts the roadmap of the deep learning algorithm selection, training, and evaluation criteria for a particular task and required libraries/tools for training the model. Figure 10 shows taxonomy of various deep learning techniques of classification [239]-[245], detection algorithms [246], unsupervised learning [247] [248], tracking [249]-[258], and trajectory prediction [259]-[266]. Since various tasks in sports such as classification/detection, tracking, and trajectory prediction show great advantages in various sports.

# 5 Available Datasets of Sports

In this section, a brief description of some sports video (or) image datasets which are available publicly with annotations are provided. Utilizing these shared datasets provides a platform for comparison of performance of algorithms with common data for improving the transparency in research in this domain. Additionally, sharing the data among the users (researchers) reduces extremely time-consuming efforts of capturing and annotating large quantities of videos in diversified areas. This allows users to get a benchmark scores for the algorithms developed.

These shared datasets can be categorized into two types: Videos or still images, which are typically taken with moving cameras, particularly of individual athletes or of team sports, for the purpose of recognition of player actions [91, 160, 224, 225, 267, 269, 270, 273], event detection and classification [32, 96, 268], which are often captured using several setups of static cameras, for the purpose of detection and tracking of player/ball [272, 274, 277], pose estimation [276], and sports event summarization [275] in the team plays. One dataset focuses on the actions of the spectators of an event in sports rather than the players. These datasets which are available for analysis are large performing a great variety of actions. Table 10 describes the available datasets of various sports, mode of dataset, annotated parameters, number of frames and length of the video.

The parameters which are annotated in ISSIA dataset relate to positions of the ball, player and referee in each video from each camera. The images shown in figure 11 are few sample frames from ISSIA dataset.

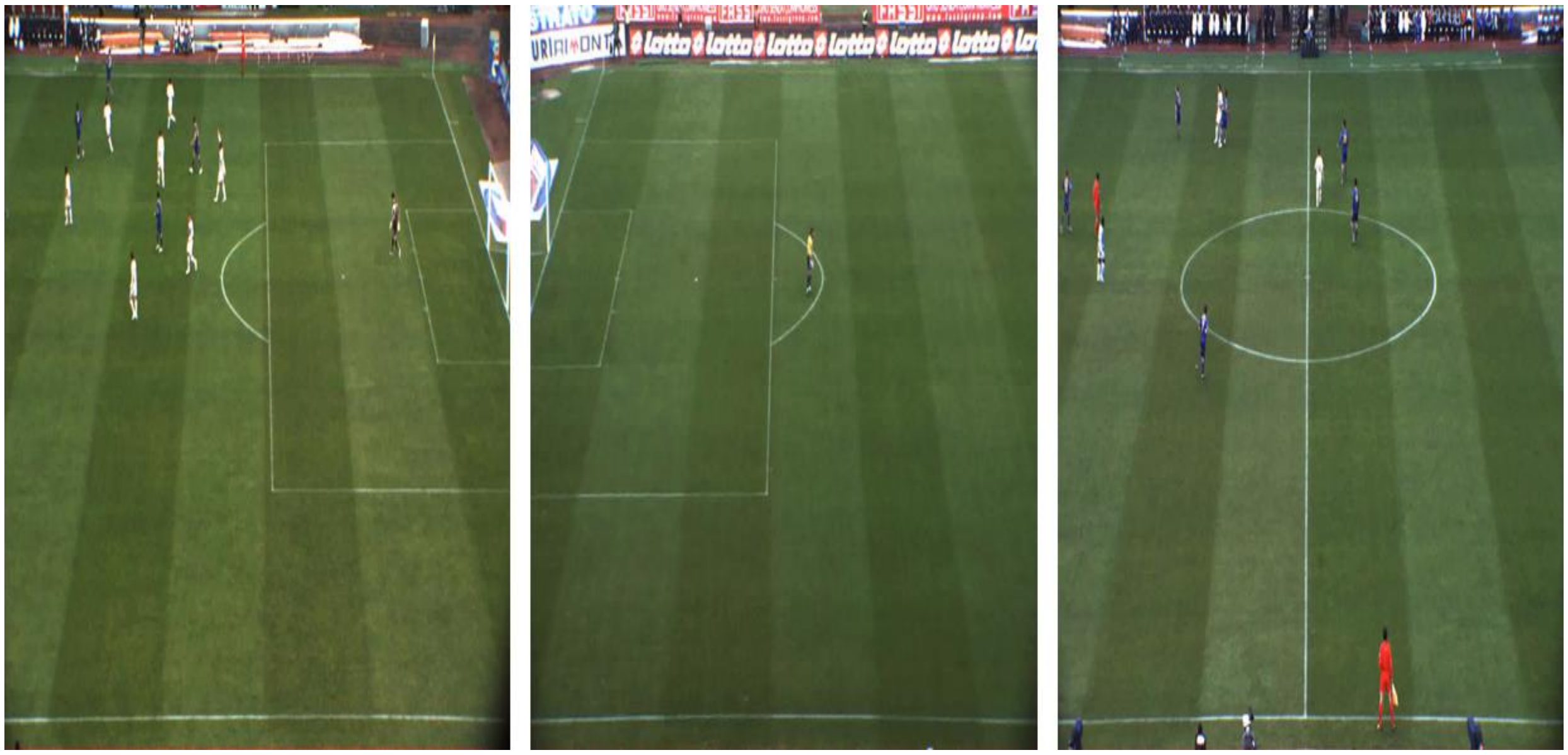

*Figure 11 Sample frames from ISSIA dataset [274]*

The parameters which are annotated in the TTNet dataset are the ball bouncing moments, ball hitting the net and empty events. The images shown in figure 12 are few samples frames from the TTNet dataset.

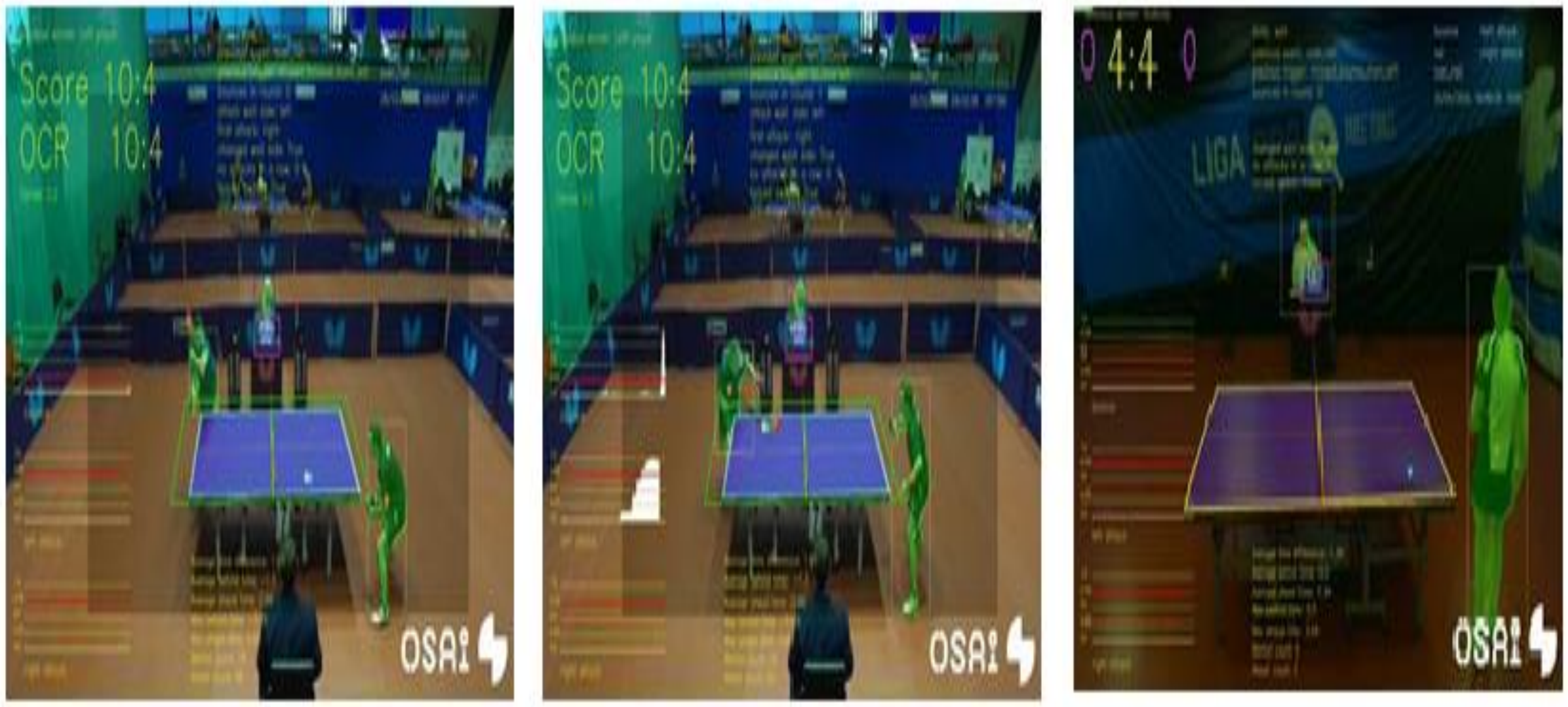

*Figure 12 Sample frames from TTNet dataset [268]*

For creation of APIDIS dataset videos are captured from seven cameras from above and around the court. The events which are annotated in this dataset are player positions, movements of referee, baskets and position of the ball. The images shown in figure 13 are the few samples from the APIDIS dataset.

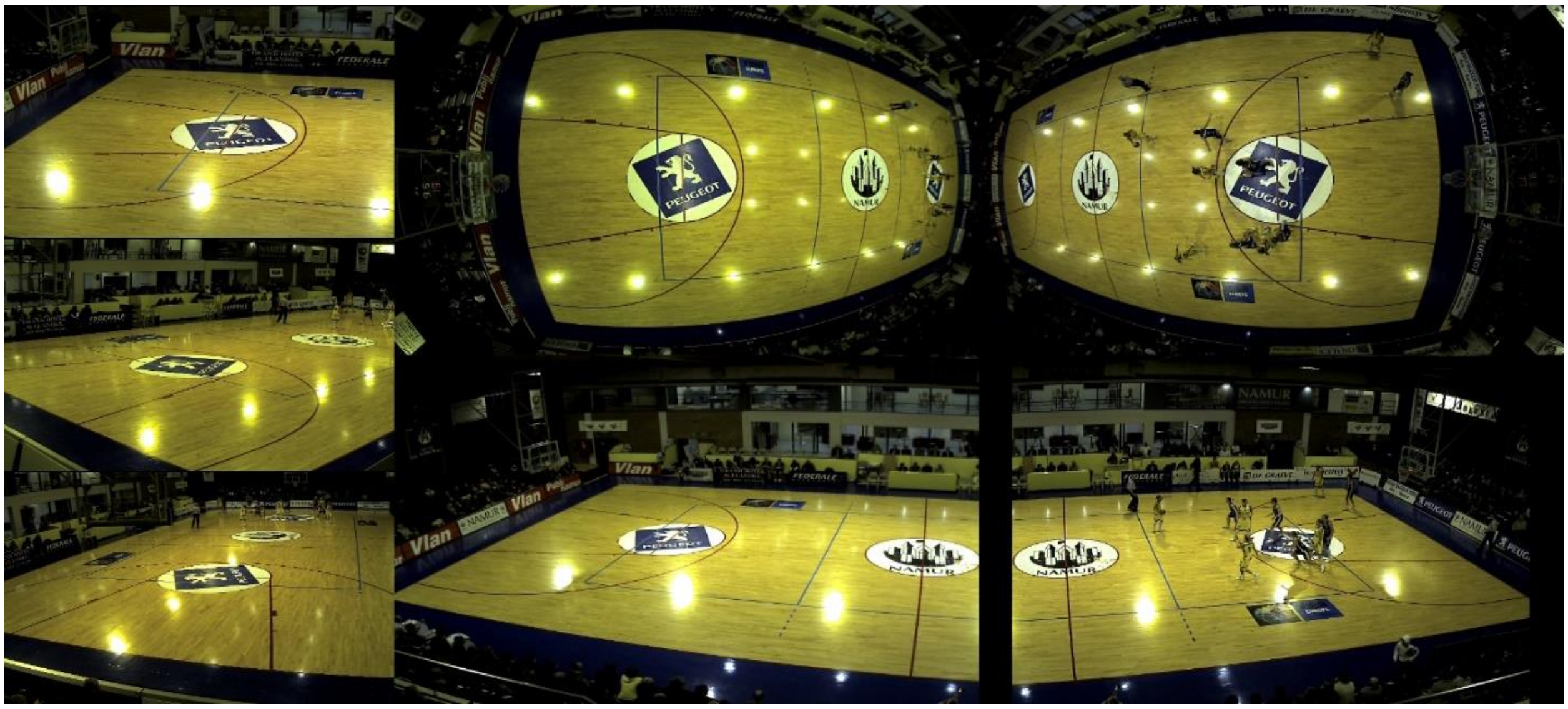

*Figure 13 Sample frames from APIDIS dataset [275]*

| Table 10 Details of available dataset | | | | | |
|---|---|---|---|---|---|
| **Ref.** | **Sport** | **Dataset** | **Mode of Dataset** | **Annotated parameters** | **Length of Video and Number of Images** |
| K Ali et al. 2021 [96] | Soccer | Image type Football Keyword Dataset | Event detection and Classification. | Events such as free kick, penalty kick, tackle, red card, yellow card. | Dataset was categorized as train, test, and validation with 5000, 500, and 500 images. |
| X Gu et al. 2020 [267] | Basketball | Basketball dataset | Action Recognition | Dribbling, Passing, Shooting | Dataset consists video of 8 hours duration, 3399 annotations and 130 samples of each class. |
| A Gupta et al. 2020 [116] | Cricket | Video type Cricket Strokes Dataset | Cricket Stroke Localization. | Annotated with strokes played in untrimmed videos. | Highlights dataset and the Generic dataset which comprised of Cricket telecast videos at 25FPS. |
| Roman V et al. [268] | Table Tennis | Video type TTNet dataset | Ball detection and Event Spotting | Ball bounces, Net hit, Empty Events | 5 Videos of 10-25 min duration for training and 7 short videos for testing |
| Silvio Giancola et al. 2018 [269, 91] | Soccer | Image type SoccerNet and SoccerNetv2 dataset | Action spotting in soccer videos | Goal, Yellow/Red Card, Substitution | Handles a length of video about 764 hours and 6,637 moments which are split into three major classes (Substitution, Goal and Yellow/Red Card). |
| Mostafa S. Ibrahim et al.2016 [160] | Volleyball | Volleyball dataset | Group activity recognition | Person-level actions, Temporal dynamics of a person’s action and Temporal evolution of group activity | 1525 frames were annotated. |
| D. Conigliaro et al. 2015 [270] | Hockey | The Spectators Hockey (S-HOCK) | Analyzing Crowds at the Stadium | spectator categorization such as position, head pose, posture and action | Video type dataset.<br>31 sec<br>30 fps |
| A. Karpathy et al. 2014 [271] | 487 classes of sports | Sports-1M dataset | Sports classification and activity recognition. | Activity labels | 5 m 36 sec |
| S.A. Pettersen et al. 2014 [272] | Soccer | Player position in soccer video Dataset | Player tracking system | Trajectories of players | 45 min |
| G. Waltner et al. 2014 [224] | Volleyball | Indoor volleyball dataset | Activity Detection and Recognition | Seven activities such as serve, reception, setting, attack, block, stand, and defense/move are | 23 min<br>25 fps |
| | | | | annotated to each player in this dataset. | |
| J.C. Niebles et al. 2010 [273] | Various jump games, various throw games, bowling, tennis serve, diving, and weightlifting. | Olympic Sports Dataset | Recognition of complex human activities in sports | Different poses in different sports | Video type dataset. It contains 16 sports classes, with 50 sequences per class. |
| V.Ramanathan et al. [32] | Basketball | NCAA Dataset | Event Recognition | Event classification, Event detection and Evaluation of attention. | The video type dataset. Length 1.5 hours long Annotated with 11 types of events. |
| T. D'Orazio et al. 2009 [274] | Soccer | ISSIA Soccer Dataset | Objective way of Ground Truth Generation | Player and Ball trajectories | 2 min 25 fps |
| C.D. Vleeschouwer et al. 2008 [275] | Basketball | APIDIS Basketball Dataset | sport-event summarization | Basketball events like position of players, referees, ball. | 16 mina, 22fps |
| W Zhang et al. [276] | Badminton, Basketball, Football, Rugby, Tennis, Volleyball | Martial Arts, Dancing and Sports dataset | 3D human pose estimation. | It is annotated with 5 types of actions. | Size of dataset is 53000 frames. |
| KhurramSoomro et al. 2014 [225] | Kicking, golf swing, lifting, diving, riding horse, skateboarding, running, walking, swing-side. | UCF Sports Action Dataset | Action Recognition | Action localization and the class label for activity recognition. | 13k clips and 27 h of video data |
| N Feng et al. 2020 [277] | Soccer | SSET | Shot segmentation, Event detection, Player Tracking | Far-view shot, Medium-view shot, Close-view shot, Out-of-field shot, and Playback shot. | 350 soccer videos of total length 282 h 25 fps |

# 6 GPU-Based Work Stations and Embedded Platforms

To find the target, GPU-constrained devices such as Raspberry Pi, Latte Panda, Odroid Xu4 and Computer Vision were used. The disadvantages of machine learning techniques are that they provide poor or inaccurate results and have issues in predicting an unknown future data. Whereas deep learning algorithms provide accurate results and also make predictions from unknown future data. Segmentation, localization and image classification are visual recognition systems which have prominent research contributions.

Among embedded AI computing platforms, NVidia Jetson devices provide low-power computing and high-performance support for artificial intelligence based visual recognition systems. Jetson modules are configured with OpenCV, cuDNN, CUDA Toolkit, L4T with LTS Linux kernel and TensorRT.
Intel Movidius Neural Compute Stick uses Intel Movidius Neural Compute SDK in GPU-Constrained devices to deploy AI algorithms.

S. Wang et al. [199] presented a high-speed stereo vision system which can track the motion of the golf ball at a speed of 360 km/h under indoor lighting conditions. They implemented the algorithm on field-programmable gate array board with advanced RISC machine CPU. P R Kamble et al. [59] implemented a deep learning approach to track the soccer ball on NVIDIA GTX1050Ti GPU. L Chen et al. [41] implemented deep learning algorithm on GTX 1080 ti GPU, based on CUDA 9.0 and Caffe to analyze the technical features in basketball video. Table 11 shows the basic comparison between GPU based devices and GPU-constraint devices and possible deep learning algorithms to implement on various devices.

**Table 11 Comparison between Jetson Modules and GPU-Constrained Devices**

| | **Jetson TX1** | **Jetson TX2** | **Jetson AGX Xavier** | **Raspberry Pi Series** | **Latte Panda** | **Odroid Xu4** |
|---|---|---|---|---|---|---|
| **CPU** | Quad core ARM Processor. | Dual core Denver CPU and quad core ARM 57 | 64 bit CPU and 8-core ARM | 63-bit quad core ARM | Intel Cherry Train quad core CPU | Cortex A7 octa core CPU |
| **GPU** | NVidia Maxwell with CUDA cores | NVidia Pascall with CUDA cores | Tensor cores + 512-core Volta GPU | - | - | - |
| **Memory** | 4GB Memory | 8 GB Memory | 16 GB Memory | 1 GB Memory | 4 GB Memory | Stacked memory of 2 Giga bytes. |
| **Storage** | 16 GB Flash Storage | 32 GB storage | 32 GB storage | Support Micro SD card | 64 GB storage | Support Micro SD Card |
| **Possible Deep Learning Algorithms to Implement.** | YOLO v2 and v3, tiny YOLO v3, SSD, Faster –RCNN and Tracking algorithm like YOLOv3+Deep SORT, YOLOv4, YOLOR. | | | YOLO, YOLO v2 and SSD-MobileNet etc. | | |

Field Programmable Gate Array (FPGA) has also been used in sports involving 3D motion capturing, object movement analysis and image recognition, etc. Table 12 describes how different researchers performed various problem statements of sports on hardware platforms such as FPGA, GPU-based devices and their results in terms of performance measures are listed.

**Table 12 Performance of various studies on hardware platforms**

| Ref | GPU Based Work Station | Embedded platform | Problem Statement | Performance measures | Result |
|---|---|---|---|---|---|
| L Wu et al. 2019 [27] | NVidia Titan X GPU. | - | Events classifications in basketball videos. | Average Accuracy | 58.10% |
| Yu Z et al. 2018 [36] | NVIDIA GTX 960 | - | Basketball trajectory prediction based on real data and generating new trajectory samples. | Measured in terms of AUC | 91.00% |
| T Liang 2020 [279] | - | FPGA | Recognizing swimming styles of a swimmer. | The result shows the three-level identification system in Average, Minimum and Maximum offset. | 4.14%, 2.16%, 5.77%. |
| J Hou et al. 2020[194] | | - | | Recall and Specificity | 85% and 96.6% |
| Y Yoon et al. 2019 [31] | NVidia GeForce GTX 1080Ti | - | Tracking ball movements and classification of players in basketball game | Precision and Recall | 74.3% and 89.8% |
| Z dou 2020 [279] | | FPGA | Ball detection and tracking to reconstruct trajectories in basketball sport. | Accuracy | >90% |
| L Chen et al. 2020 [41] | NVidia GTX 1080 ti GPU | - | Analyzing behaviour of the player. | Accuracy | 83% |
| L Yin et al. [280] | - | FPGA | Detecting movement of ball in basketball sport. | Average rate vs Frame range | Varies from 12% - 100% for different frame range. |
| R L Castro et al. 2019 [212] | NVidia GTX 1080Ti | - | Individual player tracking in sports events. | Achieved an Area Under Curve (AUC) | 66% |
| Y Ganesh et al. 2019 [83] | NVidia GeForce GTX 1080Ti | - | Action recognition in soccer | Accuracy in terms of F1 Score | 52.80% |
| H Bao et al. 2020[281] | - | FPGA | Movement classification in basketball sport based on Virtual Reality Technology to improve basketball coaching. | Accuracy | 93.50% |
| P.R Kamble et al. 2019 [59] | NVidia GTX 1050Ti | - | Detection and tracking of ball in soccer videos. | Accuracy | 87.45% |
| G junjun [282] | | FPGA | Action recognition based on arm movement in basketball sport. | Accuracy | 92.30% |

# 7 Applications in Sports Vision

A fan who is digitally connected becomes the biggest influencer online of sports venues. The teams and stadium owners who provide plenty of personalized experiences through their custom apps, mobile phone support for the content with offers, and live updates of the game information using digital boards increase the engagement of fans and in turn generate opportunities for new revenues [283]. Figure 14 depicts where AI technology can be used within the sporting landscape.

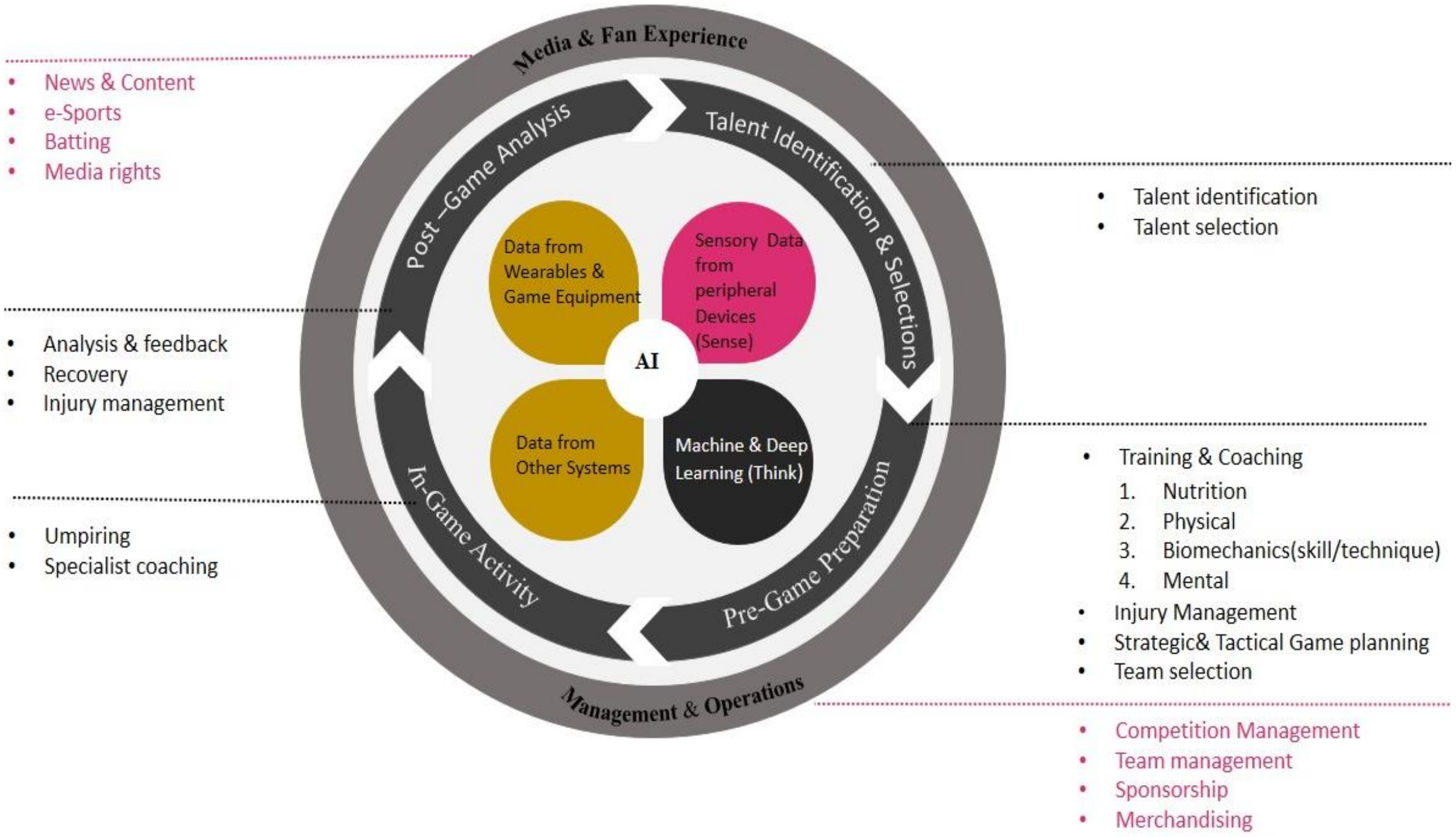

*Figure 14 AI Technology Framework for the Sport Industry*

## 7.1 Chabot's& Smart Assistants

Recently, sports teams like NHL and NBA started using virtual assistants for responding to enquires done by fans in a wide range of topics like ticketing, arena logistics, parking and other game related information. If the bots are not capable, such scenarios are handled by human intervention and they maintain customer service for that.

## 7.2 Video highlights

The challenges facing the industry include not just creation of content but also delivering it to customers through multiple devices and screens for viewing different content at different times. There is a serious demand from fans for in-depth analysis and also for commentary. Many others like action packed highlights and some behind-the -scenes content as well.

Introducing AI enables solving challenging tasks in various sports and it provides an exciting viewing experience to the audience, attracting more viewers.

## 7.3 Training & Coaching

An effective way for improving the analysis of performance of athletes and also assisting coaches with team guidance to gauge the tactics of the opponents are gaining popularity.

An application which uses AI contains huge data set of game performances and training related information which is backed up with the knowledge of several coaches and sports scientists. They act as an accumulated source for dissemination of current knowledge on dissemination of latest techniques, tactics or knowledge for professional coaches.

With the evolution of any body of knowledge on any tactic or technique, the knowledge base of AI is updated. The accumulated data can be used for training and educating sports coaches, scientists and also athletes, which in-turn leads to improved performance.

## 7.4 Virtual Umpires

In cricket and tennis, we see Video Assistant referee (VAR) and Decision Review System (DRS) which take into consideration Hawk-eye, slow motion replays and some other technologies already being used. But the catch is that these involve request from players or team for review when an umpire's or referee's decision has some uncertainty involving other parties to assist the main umpire, the whole process is also time consuming taking away the momentum and excitement of the game.

Latest camera technology supporting AI software creates a situation where an umpire's role is limited to on-field behaviour management of players rather than taking critical decisions. For example, in the case of tennis with the use of computer vision for detecting placement and speed of the ball, the need for a line umpire is completely eliminated. The future scope can be an umpire's earpiece and glasses assisting the decision instantly, eliminating the necessity of reviews.

## 7.5 AI Assistant Coaches

AI can be way more capable in situations involving a dynamic planning and analysis of the scene where a coach would rely on previous data and experience, and it cannot really be so effective to frame dynamically changing strategies in comparison to a machine. A future can be imagined in which a machine with AI running alongside the gameplay is dynamically predicting and creating strategies, helping the teams to get an edge over others. The evolution of chess technology is shown in figure 15.

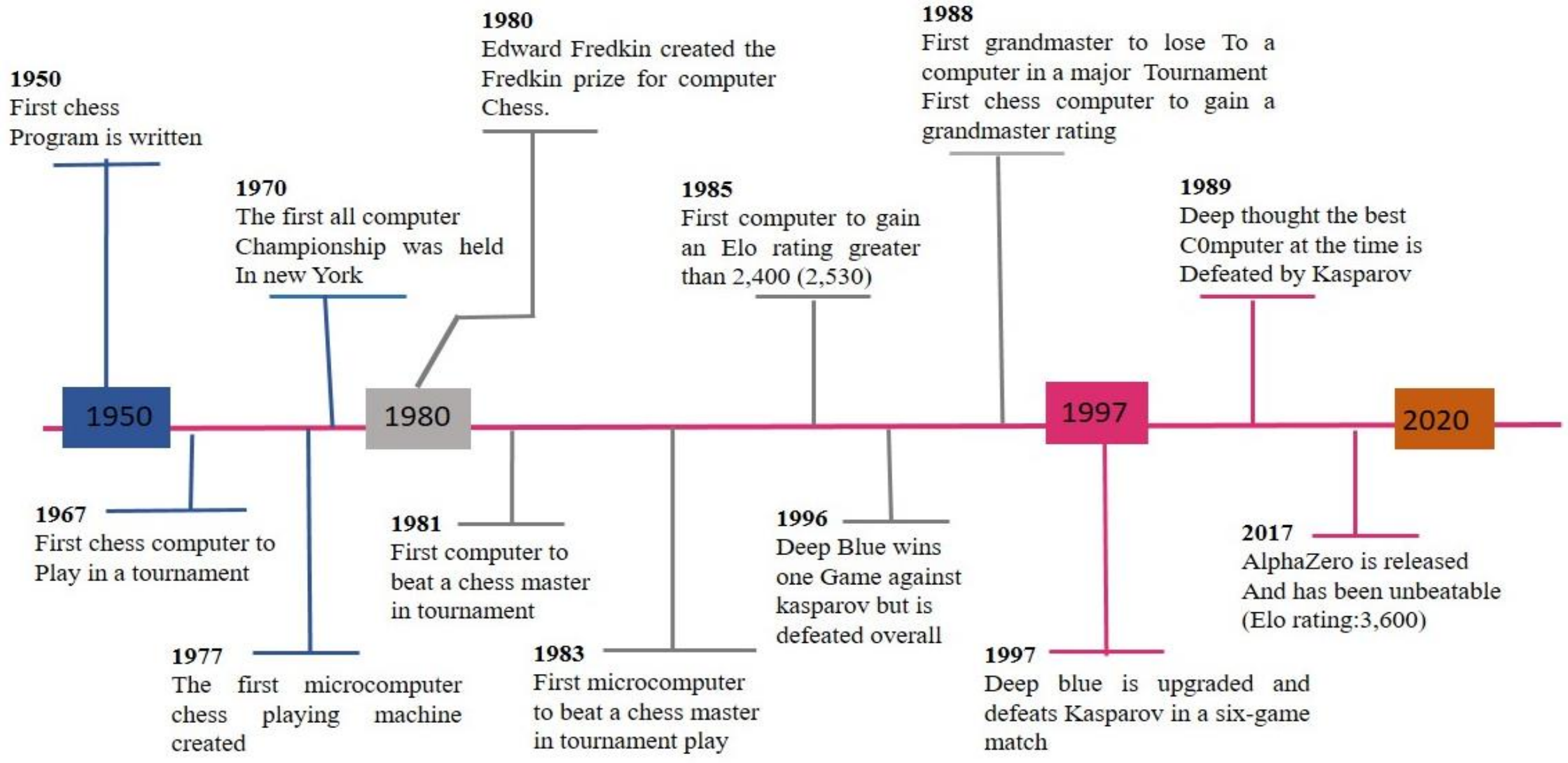

*Figure 15 Evolution of Chess technology demonstrates the speed of AI adoption.*

One example where we can see the levels AI has achieved is in Chess. The Russian Garry Kasparov who was considered world's No.1 for about 19 years with Elo rating (skill level measurement) 2851 was surpassed by Magnus Carlsen with Elo rating 2882 in 2014.

In computers, Deep Blue's rating which was 2700+ was surpassed by Deep Mind's Alpha Zero with an estimated Elo 3600, which was developed by Google's sibling DeepMind. It is developed by a reinforcement learning technique called self-play. It took just 24 hours to achieve it, proving the capabilities of the machine.

## 7.6 Available Commercial Systems for Player and Ball Tracking

Hawkeye [284, 285, 286] is the technology which is available for ball tracking in cricket, tennis and soccer. The area of primary application is officiating in tennis, cricket and soccer to enhance broadcast videos. Figure 16 shows visualization performance of the commercial systems.

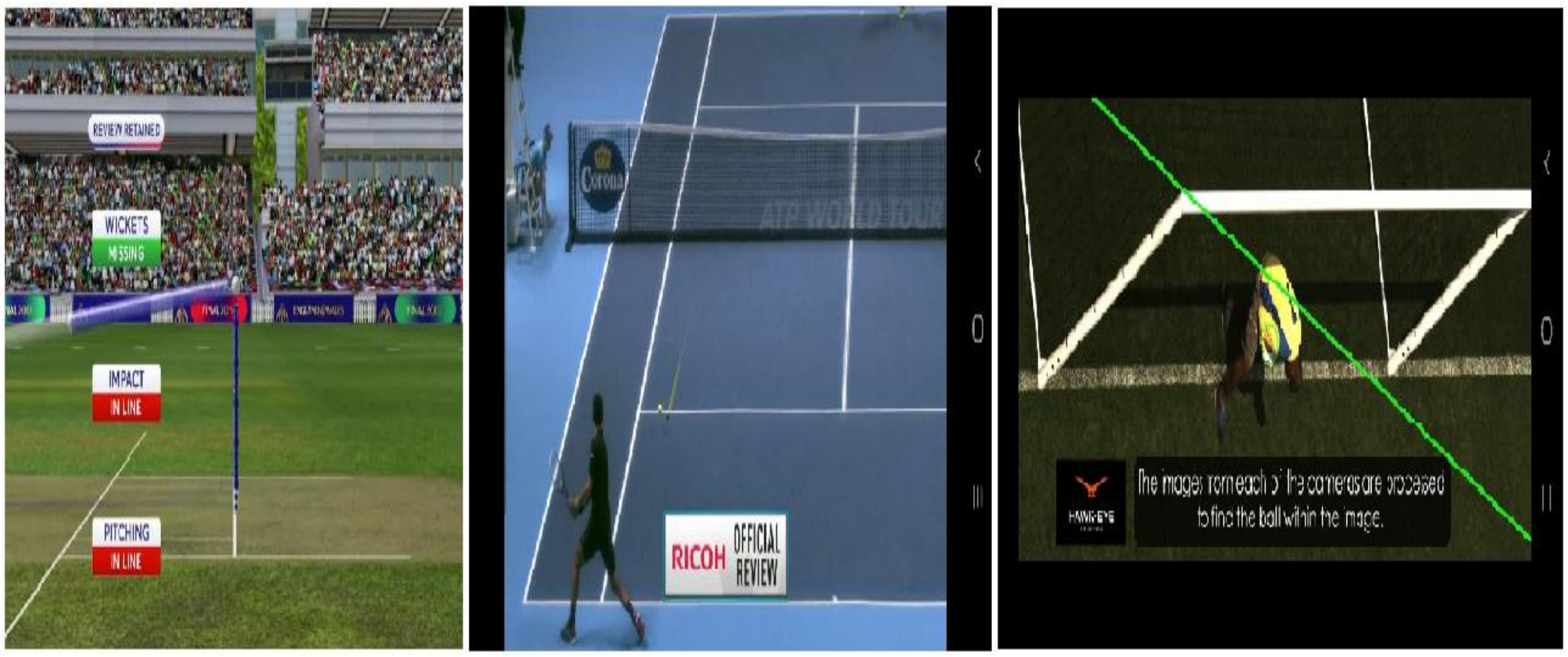

*Figure 16 Hawkeye Technology in Cricket, Tennis and Soccer [284, 285, 286]*

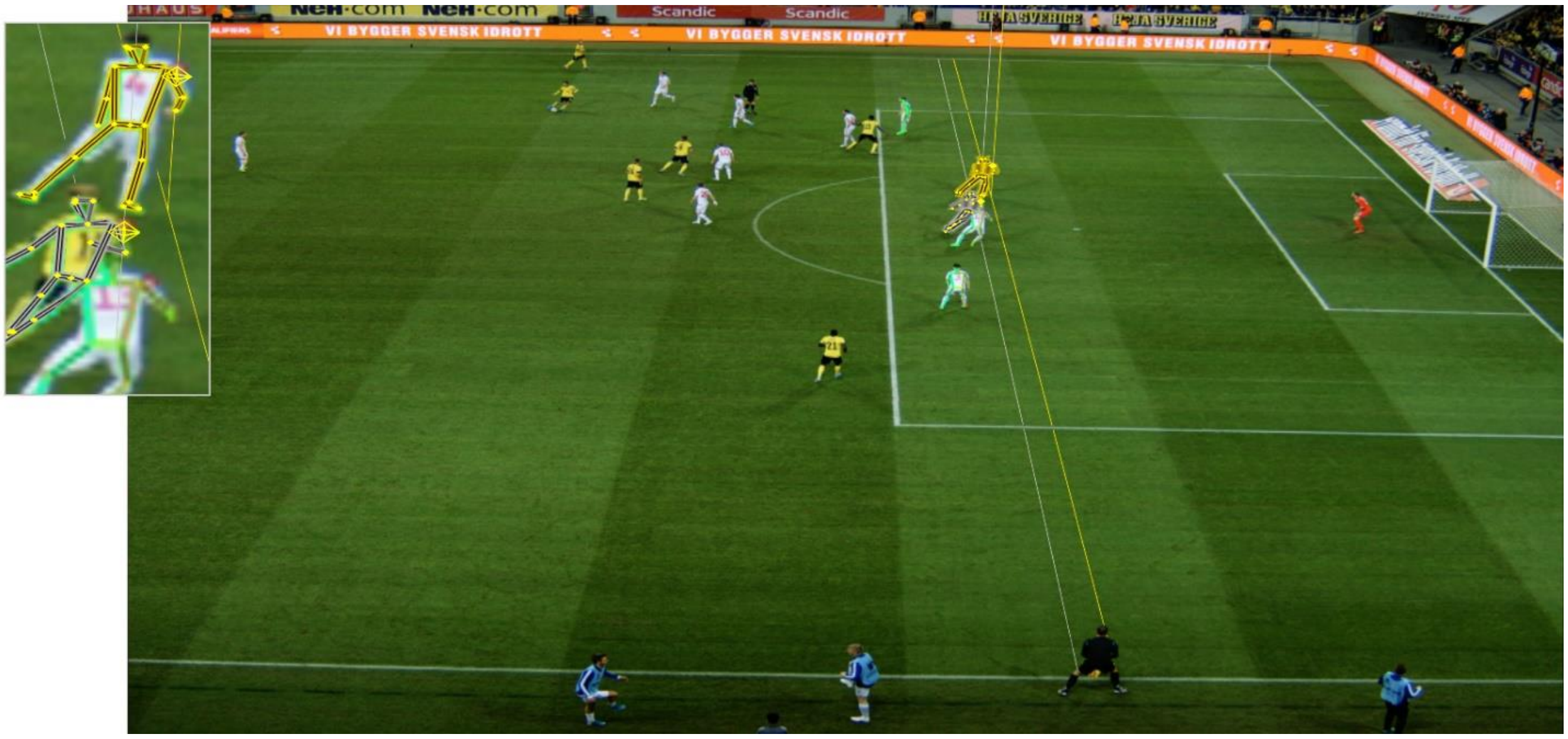

*Figure 17 TRACAB Gen5 Technology for Player Tracking [288]*

STATS SportVU [287] and ChyronHego TRACAB [288] are the technologies available for player tracking in sports. The area of primary application is to track players in various sports to analyse the performance of players and to assist coaches for training. Figure 17 shows player position and pose estimation using commercial systems. SportsVu is a computer vision technology which provides real-time optical tracking in various sports. It provides in-depth performance of any team, in terms of tracking every player from both the team to provide comprehensive match coverage, collecting data to provide tactical analysis of the match and highlighting the performance deviations to reduce injuries in the game.

## 8 Research Directions in Sports Vision

Based on the investigation of available articles in sports, we were able to come out with various research topics and identifies research directions to be taken for further research in sports. They are categorized based on the task specifics in sports applications (such as major sports in which player/ball/referee detection and tracking, pose estimation, trajectory prediction is required) as shown in figure 18 to provide promising and potential research directions for future computer vision/video processing in various sports.

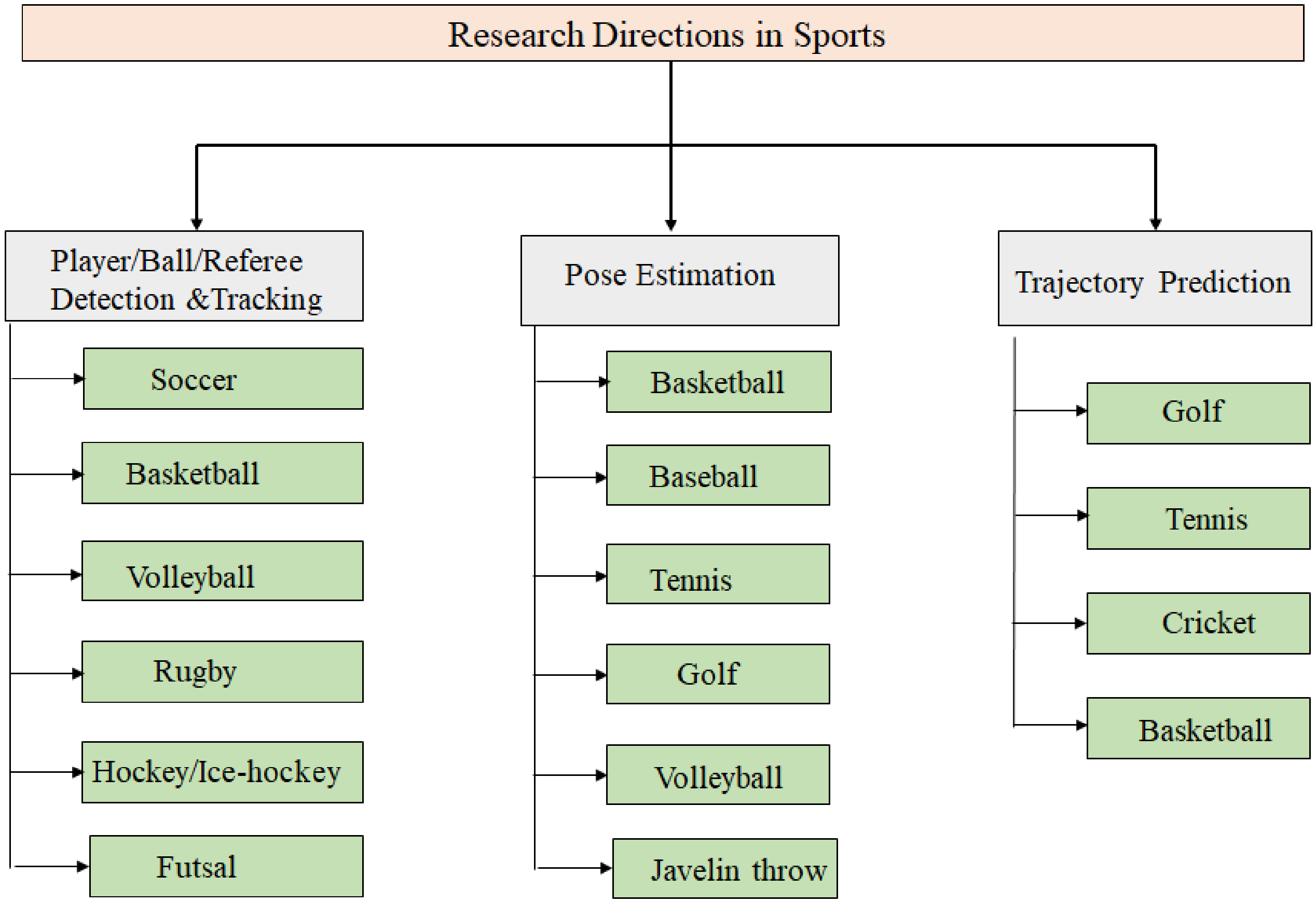

*Figure 18 Major task-specifics in sports applications*

As sporting activities are dynamic in nature, the accuracy and reliability of a single player or multi-player tracking [60, 212] in real time sports video can be enhanced by proposing a framework that learns object identities with deep representations which resolve the problem of identity switch among players [209]. By considering the temporal information, the performance of tracking algorithm can become robust to overcome problems like severe occlusions, miss-detection etc.

The accuracy of classifying different defensive strategies of various sports can be improved by labelling large spatio-temporal datasets; and, also by classifying the actions into subtypes [83], the accuracy of action recognition can be enhanced. The performance measures of team tactics analysis [88, 90] of soccer videos can be enhanced by analyzing player trajectories. By incorporating the temporal information, the classification accuracy can be improved while it also offers more specific insights to situations like pass events in the case of non-obvious insights in sports videos. Accurate pose detection as shown in figure 19 is still a major challenge to identify whether the player is running, jumping or walking as shown in figure 20, and also to handle the severe occlusions or identity switch among players.

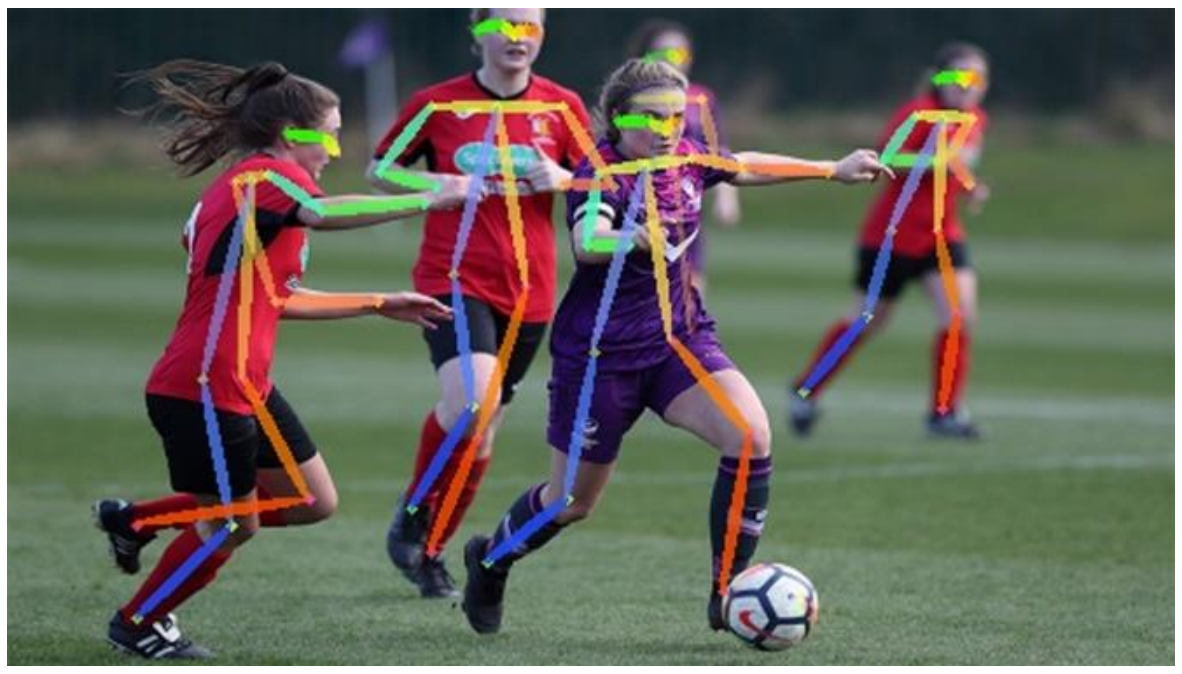

*Figure 19 Detecting body pose and limbs*

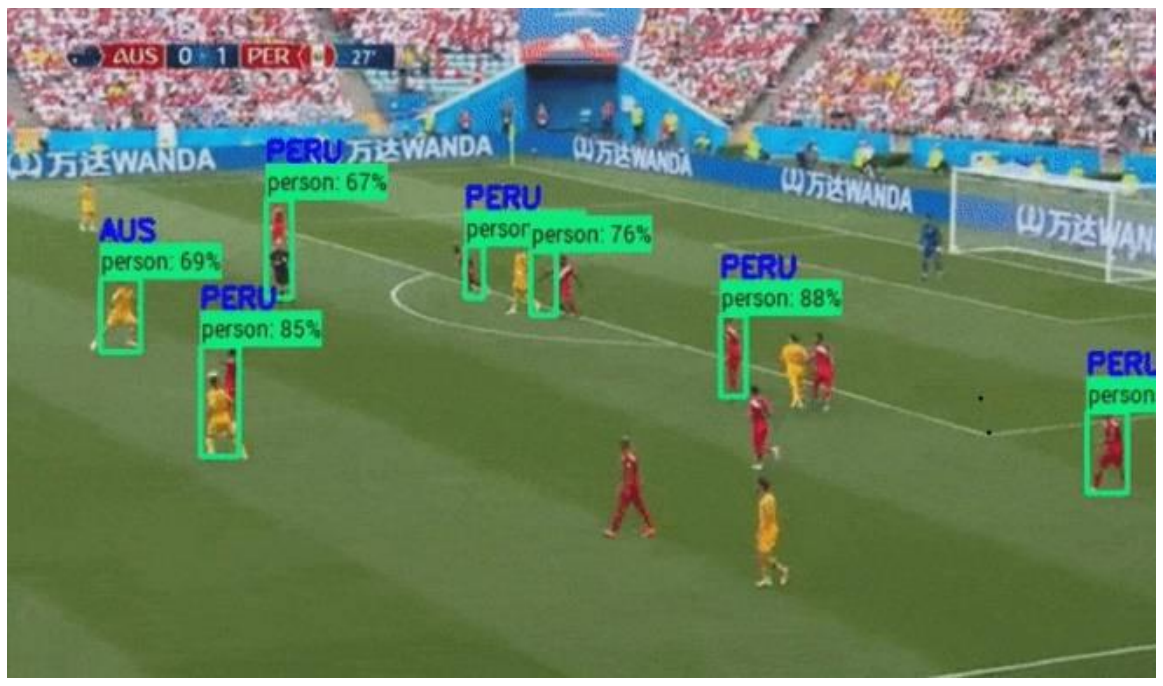

*Figure 20 Handling severe occlusions among the players*

In order to assess the player's batting caliber in cricket, certain aspects of batting also need to be considered i.e. position of batsman before playing shot and way of batting shots for a particular bowling type needs to be modeled [111]. Classification technique can minimize false positives and false negatives to detect and classify umpire poses [127]. Detecting various moments like whether the ball hit the bat and precise detection of player and wicket skipper at the moment of run outs, as shown in figure 21, is still a major issue in cricket. Predicting the trajectory of ball bowled by spin bowlers as shown in figure 22 can be resolved accurately by labeling large dataset and modeling using SOTA algorithms.

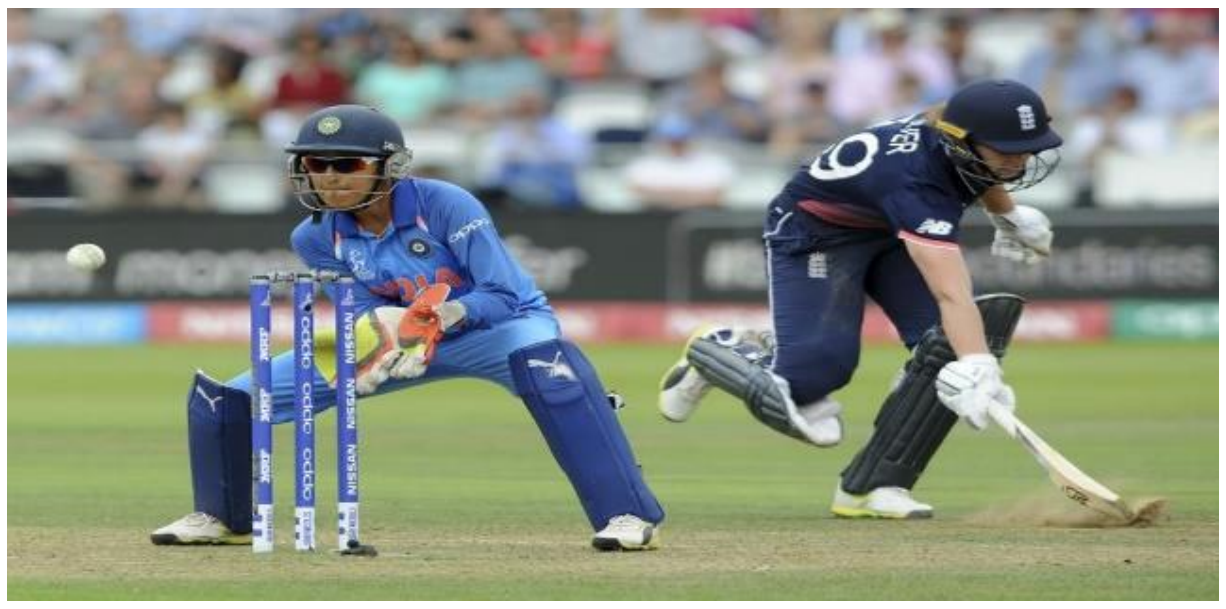

*Figure 21 Precise detecting at the moment of run outs*

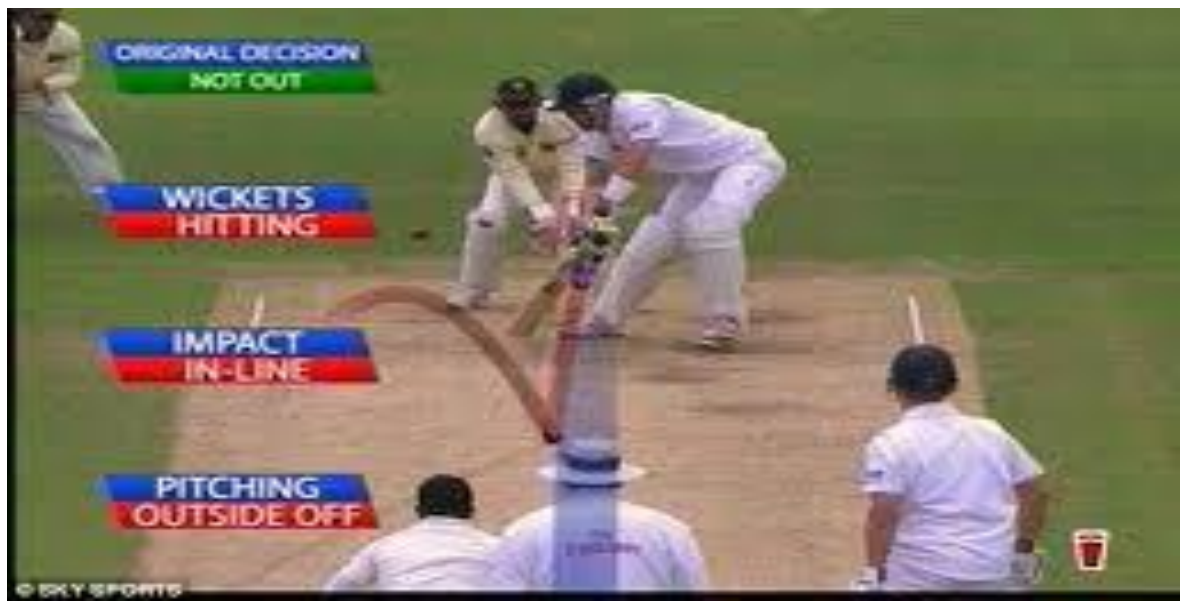

*Figure 22 Predicting trajectory of ball, in-line or out-line etc.*

Recognition accuracy of player action in badminton game [187] can be improved by SOTA computer vision algorithms and fine tuning end-to-end manner with a larger dataset on feature extracted at different fully connected layers. In the implementation of automatic linesman system in badminton game, the algorithm is not robust to the far views of the camera, where illumination conditions heavily impact the system while the speed of the shuttle cock is also a major factor for poor accuracy. So, it is needed to track the path, which becomes simpler to referee to decide if a shuttle lands out or in as shown in fig 23.

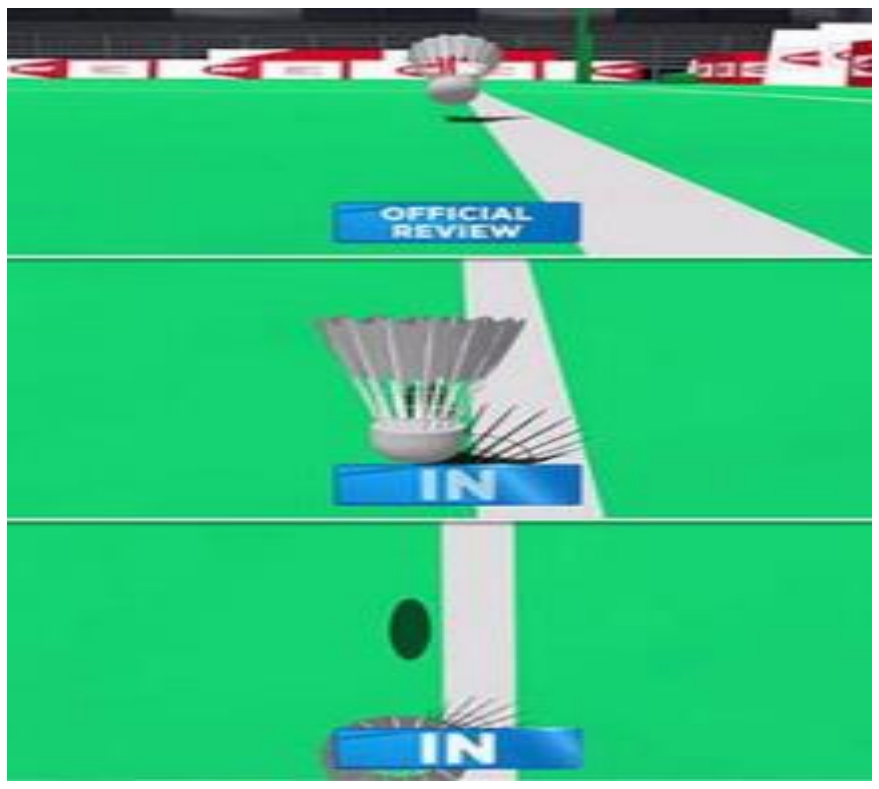

*Figure 23 Exact spot on which shuttle lands*

## 8.1 Open Issues and Future Research Areas

Computer vision plays a vital role in the area of sports video processing. To analyse sports events, there are many issues open for research. Calibration and viewpoints of the camera to capture the sports events such as close-up views, far views and wide views in degree of occlusions etc., are still issues that have not been satisfactorily addressed.

Detecting the ball in various sports helps to detect and classify various ball based events such as goals, possession of the ball and many other events. Due to the size, speed, velocity and unstructured motion of the ball compared to players and playfield in various sports, it is still an open issue to detect and track the ball. Various AI algorithms have been developed to achieve better performance in various sports such as soccer, basketball, tennis and badminton in terms of detecting and tracking with respect to various aspects of the ball.

Tracking players and ball is the one of the most open area for research which includes various issues such as fast and frequent movements of the players, similar appearance of players due to jersey colour in team sports, often partial and full occlusions of players, etc. Various algorithms use linear motions for multi-player tracking, resulting in poor performance but solves data association problems with appearance models. However, this algorithms fails in various conditions such as severe occlusions, ambiguity of appearance between players etc.

## 8.2 Future Research Trends according to Methodologies in Sports Vision

In this section, we aim to set forth the methodological approach to various components of detection, classification and tracking in sports. By considering the deep analysis of sports studies, it will be clear that the performance of the algorithm depends on the type (annotation parameters) of dataset used, which is carried out based on loss functions and evaluation metrics. The major difficulties in real time use of AI algorithms in various research areas of sports are accuracy, computation speed, size of the model etc. Considering all these aspects, the development of future trends based on contemporary ideas are presented below.

- Due to the continuous movements of player, jersey number encounter serious deformation and various image size and low resolution makes difficult to read jersey number [201]. Player's similar appearance and severe occlusions make difficult to track and identify players, referees and goal keepers reliably, which causes the critical problem of identity switch among players [209]. To solve these challenges, a framework is needed to be proposed that learns object identities with deep representations and improves tracking using information pertaining to identity.
- The algorithms employed to detect and track the state of the ball such as whether it is controlled by a player (Dribbling), moving on the ground (passed from one player to another player) or flying in the air to categorize the movement as rolling pass or lobbing pass are not robust with respect to size of ball, shape of the ball, velocity etc. and under different environmental conditions. A few researchers have come forward with novel ideas to deal with the above mentioned aspects [59, 199, 102, 207, 193, 66, 282] but research is still at a nascent stage.
- Conventional architectures of detecting, classifying and tracking are replaced with more promising and potential modern learning paradigms such as Online Learners and Extreme Learning Machine etc.

## 8.3 Different Challenges to Overcome in Sports Studies

- Classification of jersey numbers in sports like soccer and basketball is quite simple [201] as they have plain jersey but in case of the sports like hockey and American football, the jerseys are massive and have sharp contours, due to which jersey number recognition is quite hard . By implementing proper bounding box techniques and digit recognition methods, better performance of jersey number recognition in every sport can be achieved.
- Action recognition in sports videos [220, 221] is explicitly a non -1inearity problem, which can be obtained by aligning feature vectors, by providing massive amount of discriminative video representations, can provide a way to capture the temporal structure of video that is not present in the dynamic image space and analyzing salient regions of frames for action recognition.

- Provisional tactical analysis related to player formation in sports such as soccer [97], basketball, rugby, American football and hockey etc., and pass prediction [84, 85, 86, 93], shot prediction [76], expectation for goals given a game state [77] or possession of ball, or more general game strategies can be achieved through AI algorithms.
- Recognition of fine-grained activity of typical badminton strokes can be performed by using off-the-shelf sensors [188], and it can be replaced with automatic detection and tagging of aspects/events in the game and use of CCTV-grade digital cameras without additional sensors.
- Identity of the player is lost when the player moves out of the frame and to retain the identity when the players reappear in subsequent frames, the player must be recognized. The key challenges for player recognition is detecting the pose of the player [54] which is the most difficult recognition challenge, especially in case of resolution effects, variable illuminations or lighting effects and severe occlusions.

# 9 Conclusion

Sports video analysis is an emerging and very dynamic field of research. This study comprehensively reviewed sports video analysis for various applications such as tracking player or ball in sports and predicting the trajectories of player or ball, player skill and team's strategies analysis, detecting and classifying objects in sports. As per the requirements of deploying computer vision techniques in various sports, we have provided some of the publicly available datasets related to a particular sport. Detailed discussion on GPU based work station, embedded platforms and AI applications in sports have been presented. We have presented various classical techniques and AI techniques employed in sports, their performance, pros, cons and suitability to particular sports. We have listed probable research directions, existing challenges and current research trends with a brief discussion and also widely used computer vision techniques in various sports.

Individual player tracking in sports is very helpful for coaches and personal trainers. Though the sports includes particularly challenging tasks like similarities between players, generation of blurry video segments in some cases, partially or fully occlusions between players, invisibility of jersey number in some cases etc., computer vision is the best possible solution to achieve.

Classification of jersey numbers in sports like soccer and basketball is quite simple as they have plain jersey but in case of sports like hockey and American football, the jerseys are massive and come with sharp contours, due to which the jersey number recognition is quite hard. By implementing proper bounding box techniques and digit recognition methods, better performance vis-a-vis jersey number recognition in every sport can be achieved. As the appearance of players varies from sport to sport, the algorithm trained on one sport may not work when it is tested on another sport. The problem may be solved by considering a dataset which contains a small set of samples from every sport for fine-tuning.

In case of multi-player tracking in real time sports videos, severe occlusions cause critical problem of identity switch among the players.  Continuous movement of player's makes difficult to read jersey number. A player's similar appearance to another and severe occlusions make it difficult to track and identify players, referees and goal keepers reliably. Multiple object tracking in sports is a key prerequisite for realization of advanced operations in sports, such as player movement and their position in sports which will give good objective criteria to the team manager for developing a new plan to improve team performance as well as evaluate each player accurately.

Commercially used multi-camera tracking systems of players rely on some mixture of manual and automated tracking and player labeling. Optical tracking systems are a good approach for tracking players occluding each other or players having a similar appearance. The algorithm may detect false positives from out of the court such as fans wearing team uniform, as the appearance of fans is similar to that of players. This can be eliminated by estimating the playground area or broadcast camera parameters with extra spatiotemporal locations of player positions.

Action recognition in sports videos is explicitly a non-linearity problem, which can be obtained by aligning feature vectors, by providing massive amount discriminative video representations; to capture the temporal structure of the video that is not present in the dynamic image space and analyzing the salient regions of the frames for action recognition.

The algorithms employed so far for detecting and tracking ball movements began with estimating 3D ball position in trajectory. Employing these methods is very critical, as they include a lot of mathematical relations and require reliable reference objects to construct the path of the trajectory. Kalman filter and particle filter based methods are robust with respect to size, shape and velocity of the ball. However, the methods fail to establish the track when the ball reappears after occlusion. Trajectory based methods solve the problem of occlusion and are robust in obtaining data with regard to missing and merging balls but fail in case of the size and shape of the ball. Data association methods are best suited for detecting and tracking small size balls in small courts like tennis but are not suited for challenges in sports like basketball, soccer, volleyball etc. AI algorithms predict the precise trajectories of the ball from a knowledge of previous frames and are immune to challenges like air friction, ball spin and other complex ball movements. A precise database which includes different size and shape of the ball has to be introduced in order to detect the ball position and enable tracking algorithms to perform efficiently.

Detection and tracking of player, ball and assistant referee as well as semantic scene understanding in computer vision applications of sports is still an open research area due to various challenges like sudden and rapid changes in movements of the players and ball, similar appearance, players with extreme aspect ratios (players will have extremely small aspect ratios in terms of height and width when they fall down on the field) and frequent occlusions. Future scope of the computer vision research in the sports therefore handles limitations more accurately on different AI algorithms.

As the betting process involves financial assets, it is important to decide which team is likely to win; therefore, bookmakers, fans and potential bidders are all interested in estimating odds of the game in advance. So, provisional tactical analysis of field sports related to player formation in sports such as soccer, basketball, rugby, American football, and hockey etc., as well as pass prediction, shot prediction, and expectations of goals in a given a game state or a possession, or more general game strategies are needed to be analyzed in advance.

Tracking algorithms which are used in various sports cannot be compared on a common scale as experiments, requirements; situations and infrastructure in every scenario differ. Determining the performance benchmark of algorithms quantitatively is quite difficult due to the unavailability of a comparable database with ground truths of different sports differing in many aspects. In addition to these, there are additional parameters like different video capturing devices and their parameter variations which lead to difficulty in building an object tracking system in the sports field.

## Abbreviations:

ANN = Artificial Neural Network

AI = Artificial intelligence

AUC = Area under Curve

BEI-CNN = Basketball Energy Image - Convolutional Neural Network

Bi-LSTM = Bi-directional Long Short Term Memory

CNN = Convolutional Neural Network

CPU = Central Processing Unit

CUDA = Compute Unified Device Architecture

DELM = Deep Extreme Learning Machine

DeepMOT = Deep Multi Object Tracking

Deep-SORT = Simple Online Real Time Tracking with Deep Association

DRS = Decision Review System

ELM = Extreme Learning Machine

Faster-RCNN = Faster-Regional with Convolutional Neural Network

FPGA = Field Programmable Gate Array

GAN = Generative Adversarial Network

GDA = Gaussian Discriminant Analysis

GPU = Graphical Processing Unit

GRU-CNN = Gated Recurrent Unit - Convolutional Neural Network

GTX = Giga Texel Shader eXtreme

HOG = Histogram of Oriented Gradients

HPN = Hierarchical Policy Network

KNN = K-Nearest Neighbor

LSTM = Long Short Term Memory

Mask R-CNN = Mask Region-based Convolutional Neural Network

NBA = National Basketball Association

NHL = National Hockey League

R-CNN = Region-based Convolutional Neural Network

ResNet = Residual neural Network

RISC = Reduced Instruction Set Computer

RNN = Recurrent Neural Networks

SOTA = State Of The Art

SSD = Single-Shot Detector

SVM = Support Vector Machine

VAR = Video Assistant Referee

VGG = Visual Geometry Group

YOLO = You Only Look Once